\documentclass[journal]{IEEEtran}

\def\endthebibliography{%
  \def\@noitemerr{\@latex@warning{Empty `thebibliography' environment}}%
  \endlist
}

\usepackage{cite}

\usepackage[pdftex]{graphicx}
\graphicspath{{./imgs/pdf/}{./imgs/jpeg/}{./imgs/png/}{./imgs/eps/}}
\DeclareGraphicsExtensions{.pdf,.jpeg,.png, eps}
\usepackage[colorlinks=true, linkcolor=blue, citecolor=blue, urlcolor=blue]{hyperref}

\usepackage{amsmath}
\usepackage{amssymb}
\usepackage{mathtools}

\usepackage{bm}
\usepackage{booktabs}

\usepackage{algorithmic}

\usepackage{array}

\usepackage[caption=false,font=footnotesize]{subfig}
\usepackage{svg}
\usepackage{stfloats}

\usepackage{url}
\usepackage{multirow}

\usepackage{lipsum}
\usepackage{siunitx}
\usepackage{amsthm}
\usepackage[nolist,nohyperlinks]{acronym}
\begin{acronym}
    \acro{DoFs}{degrees of freedom}
    \acro{1D}{one-dimensional}
    \acro{3D}{three-dimensional}
    \acro{PCC}{piecewise constant curvature}
    \acro{PD}{proportional-derivative}
    \acro{FEM}{finite element method}
    \acro{MOR}{model order reduction}
    \acro{ROM}{reduced order model}
    \acrodefplural{ROM}{reduced order models}

    \acro{EoM}{equation of motion}
    \acrodefplural{EoM}{equations of motion}
    \acro{PDE}{partial differential equation}
    \acrodefplural{PDE}{partial differential equations}
    \acro{ODE}{ordinary differential equation}
    \acro{ODEs}{ordinary differential equations}
    \acro{BVP}{boundary value problem}

    \acro{CRT}{Cosserat rod theory}
    \acro{KRT}{Kirchhoff rod theory}
    \acro{EBRT}{Euler-Bernoulli rod theory}
    \acro{PCS}{piecewise constant strain}
    \acro{GVS}{Geometrical Variable Strain}
    \acro{ISP}{Implicit Strain Parameterization}
    \acro{DER}{Discrete Elastic Rod}
    \acro{POD}{Proper Orthogonal Decomposition}
    \acro{SVD}{Singular Value Decomposition}
    
    \acro{CSM}{continuum soft manipulator}           
    \acro{PCM}{parallel continuum manipulator}
    \acro{SMA}{shape memory alloy}
    \acro{TCP}{twisted and coiled polymer}
    \acro{CTR}{concentric tube robot}
    \acro{ETR}{eccentric tube robot}
    \acrodefplural{CSM}{continuum soft manipulators} 
    \acrodefplural{PCM}{parallel continuum manipulators}
    \acrodefplural{SMA}{shape memory alloys}
    \acrodefplural{TCP}{twisted and coiled polymers}
    \acrodefplural{CTR}{concentric tube robots}
    \acrodefplural{ETR}{eccentric tube robots}

    \acro{MB}{model-based}
    \acro{MP}{Multi-Point}
    \acro{TT}{trajectory tracking}
    \acro{IK}{Inverse Kinematics}
    \acro{ID}{Inverse Dynamics}
    \acro{MPC}{Model Predictive Control}
    \acro{SMC}{Sliding Mode Control}
    \acro{ESC}{Energy-Shaping Control}
    \acro{FL}{Feedback Linearization}
    
    \acro{PH}{Pythagorean Hodograph}

    \acro{ML}{Machine Learning}
    \acro{RL}{Reinforcement Learning}
    \acro{SL}{Supervised Learning}
    \acro{ANN}{artificial neural network}
    \acro{RNN}{recurrent neural network}
    \acrodefplural{ANN}{artificial neural networks}
    \acrodefplural{RNN}{recurrent neural networks}
    \acro{MLP}{multi layer perceptron}
    \acrodefplural{MLP}{multi layer perceptrons}
    \acro{DQL}{deep Q-learning}
    \acro{DDPG}{deep deterministic policy gradient}
    \acro{PPO}{proximal policy optimization}
    \acro{DR}{domain randomization}

    \acro{MIS}{minimally invasive surgery}
\end{acronym}

\usepackage{textalpha} 
\usepackage{tabularx}

\usepackage{moreverb,url}
\usepackage{colortbl}

\usepackage{amssymb}
\usepackage{balance}
\usepackage{pifont}
\newcommand{\cmark}{\ding{51}}
\newcommand{\xmark}{\ding{55}}

\begin{document}
\bstctlcite{IEEEexample:BSTcontrol}

\title{Rod models in continuum and soft robot control: a review}
%
%
%

\author{Carlo~Alessi$^{1, \dagger, *}$, \and 
        Camilla~Agabiti$^{2, 3, *}$, \and 
        Daniele~Caradonna$^{2, 3, *}$, \and  
        Cecilia~Laschi$^{4}$, \and 
        Federico~Renda$^{5}$, \and 
        Egidio~Falotico$^{2, 3, \dagger}$
\thanks{
$^{\dagger}$ Corresponding author.
$^{*}$ Authors contributed equally.
$^{1}$ Istituto Italiano di Tecnologia, 16163 Genoa, Italy.
$^{2}$ The BioRobotics Institute, Scuola Superiore Sant'Anna, Pisa, Italy.
$^{3}$ Department of Excellence in Robotics and AI, Scuola Superiore Sant'Anna, Pisa, Italy.
$^{4}$ Advanced Robotics Centre and Department of Mechanical Engineering, National University of Singapore, Singapore, Singapore.
$^{5}$ Khalifa University Center for Autonomous Robotics System and the Department of Mechanical \& Nuclear Engineering, Khalifa University of Science \& Technology, Abu Dhabi 127788, United Arab Emirates.
{\footnotesize \tt carlo.alessi@iit.it}
{\footnotesize \tt egidio.falotico@santannapisa.it}
}
}

\maketitle

\begin{abstract}
    Continuum and soft robots can transform automation tasks requiring compliant interaction in constrained or unstructured environments, including healthcare, agriculture, marine, and space applications. However, their complex mechanics introduce significant challenges in modeling and control. Low-dimensional continuum mechanical models, such as rod theories, effectively capture the large deformations of slender bodies in contact-rich scenarios while balancing accuracy and computational efficiency. This paper presents a vertical survey of rod models for continuum and soft robots, spanning their mathematical foundations, robot modeling, and control applications. We review the main rod theories adopted in soft robotics and introduce a deformation-based classification of rod models for continuum and soft robots. Furthermore, we survey recent model-based and learning-based control strategies leveraging rod models, highlighting their role in manipulation and physical interaction tasks. Finally, we discuss advantages, limitations, research gaps, and emerging directions of rod-based approaches. This paper aims to serve as a reference for developing models and control strategies for continuum and soft robots.
\end{abstract}

\def\IEEEkeywordsname{Note to Practitioners}

\begin{IEEEkeywords}
This paper is motivated by the challenge of modeling and controlling continuum and soft robots for compliant interaction in complex environments. To balance physical accuracy and computational efficiency, the paper focuses on rod theories, which effectively describe the large deformations of slender structures. We review the literature on rod theories in soft robotics, covering mathematical foundations, continuum and soft robot models, and control strategies based on model-based and deep learning approaches. Finally, the paper discusses the main advantages, limitations, and open challenges of current rod-based methods, outlining future directions for automation and contact-rich manipulation tasks. The presented perspectives can support the development of more reliable and deployable continuum and soft robotic systems.
\end{IEEEkeywords}

\def\IEEEkeywordsname{Index Terms}


\section{Introduction}
\label{sec:introduction}
Continuum robots are robots with distributed deformations along their structure, resulting in an infinite number of \ac{DoFs}. This property creates a hyper-redundant configuration space, allowing the robot tip to reach any point in the \ac{3D} workspace with virtually infinite configurations \cite{trivedi2008SoftRob}. Soft robots, built with soft materials or deformable structures \cite{laschi2014soft}, are a subset of continuum robots and can assume various morphologies. In this paper, we refer to both classes, considering their partial overlap. 

\begin{figure}[!t]
    \centering
    \includegraphics[width=1\columnwidth]{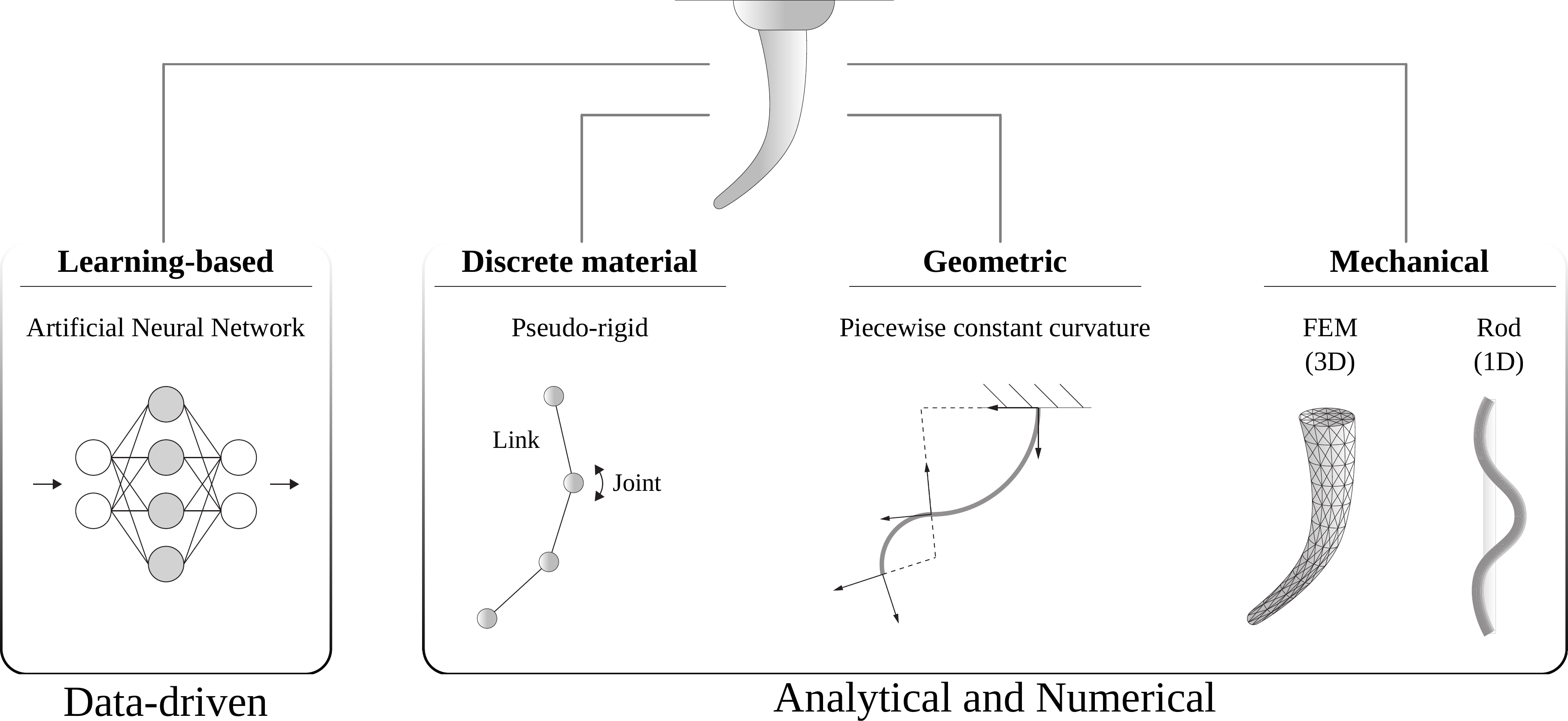}
    \caption{Overview of modeling techniques for continuum and soft robots (Sec.~\ref{sec:modeling_techniques}). This review focuses on rod models, from their mathematical formulation to control applications.}
    \label{fig:soft_robot_modelling}
\end{figure}

Continuum and soft robots can deform due to the inherent compliance of soft materials \cite{laschi2016soft}, making them promising for automation tasks requiring safe, dexterous interaction in constrained or unstructured environments. Advances in design and manufacturing technologies are accelerating the development of highly dexterous robotic systems \cite{rus2015design, WangYanmei2024Advancements}. Despite this progress, modeling and controlling soft robots remain challenging. Indeed, soft materials may introduce extreme hyper-redundancy and complex nonlinear behaviors, including hysteresis and stress softening. One of the earliest continuum formulations for hyper-redundant robots represented a manipulator as a smooth backbone curve with distributed parameters and derived \acp{PDE} for its dynamics \cite{chirikjian1993continuum,chirikjian1994hyper}. Since then, many researchers have developed diverse modeling techniques characterized by different assumptions, mathematical frameworks, and trade-offs between accuracy and computational cost \cite{armanini2023soft, gilbert2021mathematical}. As summarized in \autoref{fig:soft_robot_modelling}, soft robotics encompasses four model classes: \textit{data-driven}, \textit{discrete}, \textit{geometrical}, and \textit{continuum mechanical}. Concurrently, innovative control strategies emerged to leverage these models \cite{george2018control, della2023model, falotico2024learning}. In this paper, we focus on the achievements of rod theories from modeling to control.

Rod theories are a fundamental framework for modeling, simulation, and control in continuum and soft robotics. Despite their extensive use, the literature still lacks a comprehensive and structured review consolidating their key contributions. To address this gap, this paper investigates the role of rod theories in continuum and soft robots from a vertical perspective. We guide the reader from the mathematical formulation of rod theories, through their application to continuum and soft robot modeling, to recent developments in model-based and learning-based control. We structure the modeling literature around deformation modes, providing both a synthesis of the state of the art and a practical reference for researchers developing new continuum and soft robotic systems.

First, we report related surveys on soft robotics that provide valuable complementary perspectives, where rod models are treated marginally (Sec.~\ref{sec:related_work}). Second, we characterize the major modeling techniques to identify their strengths and limitations with respect to rod models (Sec.~\ref{sec:modeling_techniques}). Then, the main contributions of this paper are as follows:

\begin{itemize}
\item \textbf{A comparative review of four major rod theories and discretization techniques}. We present the \textit{Cosserat-Reissner}, \textit{Kirchhoff-Love}, \textit{Timoshenko}, and \textit{Euler-Bernoulli} rod models with a consistent mathematical framework, which facilitates comparison and highlight connections to alternative approaches (Sec.~\ref{sec:theoretical_background}). We consider the modeling of actuation-induced distributed loads for different actuator routings (Sec.~\ref{sec:actuation}).

\item \textbf{A structured review of rod-based models for continuum and soft robots.} We organize rod models into nine \textit{deformation classes}, providing a novel reference for the further development of soft robot models (Sec.~\ref{sec:models}). Deformation modes, such as bending and stretching, are fundamental behaviors that collectively define how soft robots move and interact with their environment. Classifying the literature by these modes highlights the mechanical effects of actuation and clarifies the associated modeling and control approaches.

\item \textbf{A comprehensive review of model-based control for \acp{CSM} using rod models.} We cover a range of approaches from foundational methods, such as inverse kinematics, inverse dynamics, and feedback linearization, to advanced techniques like model predictive control and optimal control (Sec.~\ref{sec:model_based_controllers}).

\item \textbf{A review of learning-based controllers leveraging rod models.} We emphasize \ac{RL} as a promising key enabler to achieve contact-rich manipulation tasks with \acp{CSM} (Sec.~\ref{sec:learning_based_controllers}).

\end{itemize}

Throughout the paper, we provide insights and critiques on the reviewed methods, consolidating the discussion on major trends, research gaps, and emerging challenges (Sec.~\ref{sec:discussion_and_emerging_challenges}). Although this review focuses on rod theories for modeling and control, the presented perspectives contribute to the broader advancement of soft robotics research. Finally, we conclude with a summary of the main lessons learned (Sec.~\ref{sec:conclusion}).

\section{Related Work}
\label{sec:related_work}
Soft robotics has been the subject of extensive surveys, reviews, and perspectives, which represent a wealth of knowledge and a growing interest in a thriving research field.

\textbf{Modeling-oriented surveys}: Early surveys on continuum robot modeling mainly focused on kinematic representations. The seminal work of \cite{webster2010design} reviews kinematic models of \ac{PCC} for continuum robots, presenting \textit{robot-specific} and \textit{robot-independent} mappings. In contrast, our work provides a comprehensive treatment of rod models, which extend \ac{PCC} formulations to scenarios involving significant interaction forces, such as manipulation tasks. 

Subsequent works progressively incorporated mechanics-based formulations. For instance, \cite{gilbert2021mathematical} presents a unified framework for modeling biomedical continuum robots, comparing Cosserat rod and general continuum formulations while discussing trade-offs between model accuracy and computational cost. With a similar emphasis on clinical applications, \cite{dupont2022continuum} reviews the modeling and control of several continuum robot architectures, including tendon-driven, multibackbone, concentric tube, magnetic, and soft designs. For a broader perspective on medical continuum robotics, we also refer the reader to the survey by \textit{Burgner-Kahrs et al.} \cite{burgner2015continuum}.

Several surveys instead adopt a broader soft robotics perspective. The review by \cite{zhang2022survey} spans the entire pipeline, highlighting key technical challenges from design and actuation to modeling and control. Likewise, the perspective article by \cite{mengaldo2022concise} discusses embodied intelligence in soft robots through a unified modeling formulation and analyzes its theoretical implications. In contrast, we bridge mathematical foundations and control-oriented applications specifically for rod theories.

Other surveys concentrate on specific computational paradigms. The review by \cite{schegg2022review} emphasizes real-time \ac{FEM}-based approaches for soft robot modeling and control, while \cite{sadati2023reduced} focuses on \ac{MOR} techniques and \acp{ROM}, discussing constitutive laws, solution strategies, and reduced representations for soft robotic systems. While these works primarily address theoretical and computational aspects, we additionally discuss experimental validation and practical implications of rod models. Finally, \cite{armanini2023soft} broadly classifies soft robot models into continuum mechanics, geometrical, discrete material, and surrogate approaches, analyzing their theoretical foundations and applicability. However, control aspects are addressed only marginally. In contrast, we investigate rod models in greater depth as a specific subset of continuum mechanics approaches, emphasizing the role of deformation modes in both modeling and control.

\textbf{Control-oriented surveys}: Regarding control, \cite{george2018control} provide a broad overview of \textit{model-based} and \textit{model-free} control strategies for \acp{CSM}, however, without concern for the modeling. Building on this, \cite{wang2021survey} provides a systematic review of \ac{ML} for continuum robot control, highlighting model-free approaches for inverse kinematics and closed-loop control, and outlining trends to handle model uncertainties with data-driven methods. Similarly, \cite{wang2022control} reviews actuation mechanisms and controls, including open-loop, closed-loop, and autonomous methods. They discuss emergent directions in the control-actuator interface, underactuation, and the adoption of artificial intelligence. Subsequently, \cite{della2023model} describe the problem of model-based control of \acp{CSM} through a model-agnostic formulation of soft robot dynamics. Then, they discuss shape control and tracking problems, exploring open challenges such as underactuation, environmental interaction, actuator dynamics, and task-space control. Finally, \cite{falotico2024learning} explore how different models influence the design and performance of learning-based controllers. In contrast, we investigate model-based and learning-based controllers with a focus on rod models.

In summary, this review complements existing surveys with limited overlap by providing a focused and control-oriented treatment of rod theories for continuum and soft robots. While most prior works present broad overviews of modeling or control approaches, we connect the mathematical foundations of rod theories to their implications for robot modeling and control. In particular, we structure the literature around deformation modes, offering a novel perspective that bridges theory, practical modeling, and recent control strategies, including both model-based and learning-based methods.

\section{Features of Soft Robot Models}
\label{sec:modeling_techniques}
The soft robotics community proposed various techniques to describe continuum and soft robots. According to \cite{armanini2023soft}, soft robotics encompasses four model classes: \textit{data-driven}, \textit{discrete}, \textit{geometrical}, and \textit{continuum mechanical} (\autoref{fig:soft_robot_modelling}). Clearly, each strategy has advantages and limitations, which depend on the purpose and objectives. Also, some approaches could partially overlap or be combined to derive \textit{hybrid} models \cite{falotico2024learning}. In the context of modeling and control, we answer the following question: \textit{What are the advantages and disadvantages of rod models with respect to other techniques?}
To this end, we provide a concise characterization of the features of the existing modeling techniques and their role in the control problem. We also report the main features in \autoref{tab:features_soft_robot_models}. 

\subsection{Machine Learning-based Models}
\ac{ML} is a powerful data-driven approach to derive the forward model of \acp{CSM} \cite{laschi2023learning}. Unlike analytical methods, which require explicit geometrical and physical descriptions of the robot mechanics, \ac{ML} only requires training \acp{ANN} on descriptive motion data to approximate the map between actuation and task space \cite{chin2020machine}. 
The advantages of this approach include independence from the robot, the computational efficiency of the forward model, and the potential implicit representation of complex physical phenomena, such as hysteresis \cite{falotico2024learning}. Indeed, learned forward models were effectively employed within control policies for tracking \cite{centurelli2022closed, pique2022controlling} and throwing \cite{bianchi2023softoss}. 
However, they might become unsuitable for complex manipulation tasks due to the challenges of collecting and labeling representative interaction data \cite{alessi2024pushing}. Other limitations include the lack of interpretability of the forward model and the tendency to overfit the training data.

\subsection{Geometrical Models}
Geometrical models describe the configuration of \acp{CSM} through a finite-dimensional parametrization of their backbone shape. Rather than deriving the robot configuration directly from continuum balance laws, these approaches assume that the deformed backbone can be represented by a predefined family of curves or shape functions, whose parameters constitute the generalized coordinates of the system \cite{armanini2023soft}. Typical examples include curvature, arc length, modal amplitudes, or spline control points. 
Within this class, the most widely adopted approach is the \ac{PCC} formulation, where the robot is approximated as a sequence of circular arcs with constant curvature and negligible torsion \cite{webster2010design}. PCC models are particularly effective for slender continuum manipulators actuated by tendons or pressure chambers arranged parallel to the backbone, since these actuation patterns naturally induce constant-curvature deformations in free space. Extensions based on variable curvature \cite{mahl2014variable}, polynomial curvature \cite{della2019control}, modal representations, and spline-based parameterizations have also been proposed to improve the trade-off between model accuracy and computational efficiency. 

Geometrical models reduce the infinite-dimensional continuum problem by directly parameterizing the robot shape with a finite set of kinematic variables. In contrast to continuum mechanics approaches, this reduction is introduced at the level of the kinematic representation, and the resulting models do not rely on the underlying continuum PDEs when deriving static or dynamic equations. Instead, the governing equations are constructed directly from the generalized coordinates chosen to represent the geometry. These variables can then be used within static or dynamic equilibrium formulations, for example, through energy-based or virtual work principles.

Due to their low-dimensional structure and intuitive geometric interpretation, these models are particularly attractive for real-time estimation and control. Indeed, geometrical representations have been successfully employed in both model-based \cite{fischer2022dynamic} and learning-based control frameworks \cite{lou2024controlling}. However, their accuracy strongly depends on the validity of the assumed shape parametrization. In highly constrained interactions or under complex distributed loading conditions, the imposed geometric representation may become insufficient to capture localized deformations, shear, torsion, or material-dependent effects unless enriched with additional theories \cite{xie2022simplified}.

\begin{table}[t]
\small\sf\centering
\caption{Features of soft robotics models (Sec.~\ref{sec:modeling_techniques}). Rod models combine physical modeling capabilities with tractable formulations for simulation and control.}
\label{tab:features_soft_robot_models}
\resizebox{1\columnwidth}{!}{
\begin{tabular}{lccccc}
\toprule
\textbf{Features} & 
\textbf{ML} & 
\textbf{Geometrical} & 
\textbf{Discrete} & 
\textbf{Rod} & 
\textbf{FEM} \\
\midrule
\cellcolor{gray!20}Computational Efficiency & \cellcolor{gray!20}\cmark  & \cellcolor{gray!20}\cmark & \cellcolor{gray!20}$\sim$ & \cellcolor{gray!20}$\sim$ & \cellcolor{gray!20}\xmark   \\
\midrule
Accurate for Complex Geometries & $\sim$  & \xmark           & $\sim$        & $\sim$   & \cmark   \\
\midrule
\cellcolor{gray!20}Usable in Real-time Control   & \cellcolor{gray!20}\cmark  & \cellcolor{gray!20}\cmark & \cellcolor{gray!20}$\sim$  & \cellcolor{gray!20}$\sim$  & \cellcolor{gray!20}\xmark   \\
\midrule
Data-driven Adaptability & \cmark & $\sim$ & $\sim$ & $\sim$ & $\sim$ \\
\midrule
\cellcolor{gray!20}Handles Large Deformations & \cellcolor{gray!20}$\sim$ & \cellcolor{gray!20}$\sim$ & \cellcolor{gray!20}$\sim$ & \cellcolor{gray!20}\cmark & \cellcolor{gray!20}\cmark \\
\midrule
Incorporates Material Models & \xmark & $\sim$ & $\sim$ & \cmark & \cmark \\
\midrule
\cellcolor{gray!20}Ease of Formulation/Implementation & \cellcolor{gray!20}\cmark & \cellcolor{gray!20}\cmark & \cellcolor{gray!20}\cmark & \cellcolor{gray!20}$\sim$ & \cellcolor{gray!20}\xmark \\
\midrule
Generalization to New Tasks & \xmark & $\sim$ & $\sim$ & \cmark & \cmark \\
\midrule
\cellcolor{gray!20}Model Interpretability & \cellcolor{gray!20}\xmark & \cellcolor{gray!20}\cmark & \cellcolor{gray!20}\cmark & \cellcolor{gray!20}\cmark & \cellcolor{gray!20}\cmark \\
\midrule
Handles Physical Interactions & $\sim$ & $\sim$ & $\sim$ & \cmark & \cmark \\                               
\bottomrule
\multicolumn{6}{l}{(\cmark) = Highly Suitable or Advantageous}\\
\multicolumn{6}{l}{($\sim$) = Moderately Suitable or Partially Effective}\\ 
\multicolumn{6}{l}{(\xmark) = Less Suitable or Challenging}\\
\end{tabular}
}
\end{table}

\subsection{Discrete Models}
The concept behind discrete models is to discretize the system at the beginning of the modeling process. The main approaches in this category are lumped-mass and pseudo-rigid.

\textbf{Lumped-mass}: The lumped-mass method approximates the \ac{CSM} as a set of lumped masses, springs, and dampers \cite{habibi2020lumped}. In this case, the configuration variables represent the spatial displacements of each lumped mass. The modularity of this approach can be adapted to represent arbitrarily complex phenomena (e.g., nonlinear friction) and morphologies (e.g., hybrid kinematic chains). On the other hand, to describe \acp{CSM} with the same fidelity of continuum mechanics, they require a high number of lumped masses that entail computational effort in addition to data-intensive system identification. Nonetheless, the adoption of this method is facilitated by open-source simulators like SoMo \cite{graule2021somo} and Titan \cite{austin2020titan}.

\textbf{Pseudo-rigid}: Conversely, the pseudo-rigid approach represents \acp{CSM} with a chain of rigid links connected by joints, finding an equivalent hyper-redundant rigid robot. Their formulation allows us to exploit the standard controllers from rigid robotics directly \cite{bruno1994robotics}. Therefore, pseudo-rigid models can provide satisfactory approximations for hyper-redundant arms \cite{venkiteswaran2019shape}. However, accurately reproducing smooth continuous deformations and large curvatures generally requires a high number of discrete joints and \ac{DoFs}, increasing model complexity and computational cost.  Therefore, pseudo-rigid methods can be less practical for modeling the continuous elastic structures found in soft robotics. Nonetheless, \cite{morimoto2021model, morimoto2022characterization} successfully adapted the MuJoCo simulator \cite{todorov2012mujoco} to describe \acp{CSM} and train learning-based controllers.

\subsection{Continuum Mechanical Models}
Finally, \textit{continuum mechanical} models apply the principles of continuum mechanics to characterize soft robots with continuous configuration spaces and define deformations in physical terms. This formulation enables the simulation of physical interactions (e.g., contact with objects or fluids) and the in-depth study of robot mechanics, particularly for large deformations. These properties make them promising towards \acp{CSM} interacting with unstructured environments. Depending on the dimensionality and approximation level, we can consider \ac{3D} models based on the \ac{FEM} and low-dimensional formulations such as rod models.

\textbf{FEM}: The \ac{FEM} is one of the most widely used numerical approaches for solving \acp{PDE}, particularly for modeling \ac{3D} elastic bodies. \ac{FEM} discretizes the soft robot volume into a \textit{mesh}, a set of interconnected \textit{finite elements} (e.g., line segments, triangles, tetrahedra). The points defining these elements are called \textit{nodes}, where physical quantities like displacement or stress are computed. Then, the continuous solution can be approximated by interpolating values between the nodes. This approach enables accurate simulation of complex, nonlinear deformations to the extent of high computational costs and an involved mathematical formulation. Indeed, the solutions provided by standard commercial software (e.g., Abaqus, Ansys, COMSOL) would impede real-time control. Therefore, they were often limited to designing and simulating soft robotic components \cite{xavier2021finite} or as a benchmark. Nonetheless, recent \ac{MOR} techniques implemented in the SOFA simulator \cite{allard2007sofa, coevoet2017software, dubied2022sim} are making \ac{3D} mechanical models more affordable even for learning-based control \cite{agabiti2023whole, menager2023toward}.

\textbf{Rod}: When the robot is slender, it can be efficiently represented using elastic rod formulations, which approximate the body as a \ac{1D} continuum. In particular, the \ac{CRT} can effectively describe all modes of deformations in \acp{CSM} \cite{renda2014dynamic, gazzola2018forward, till2019real}. The \ac{CRT} associates a reference system to each cross-section along the \ac{CSM}. Each cross-section can translate and rotate relative to the adjacent cross-sections to simulate bending, twisting, shearing, and stretching. Thanks to the low-dimensional yet physics-based formulation, rod models retain most of the benefits of continuum mechanics at a fraction of the computational cost of \ac{FEM}. A limitation of rod models is their underlying assumption of structural slenderness, although heterogeneous rod assemblies can be used to represent more complex systems \cite{renda2018unified, zhang2019modeling}. Beyond modeling, rod models are also gaining popularity in model-based control \cite{caradonna2024model} and learning-based manipulation \cite{alessi2024pushing}. For these reasons, rod theories constitute the main focus of the remainder of this paper.

\section{Background on Rod Theories}
\label{sec:theoretical_background}
\begin{figure}[t]
    \centering
    \includegraphics[width=1\columnwidth]{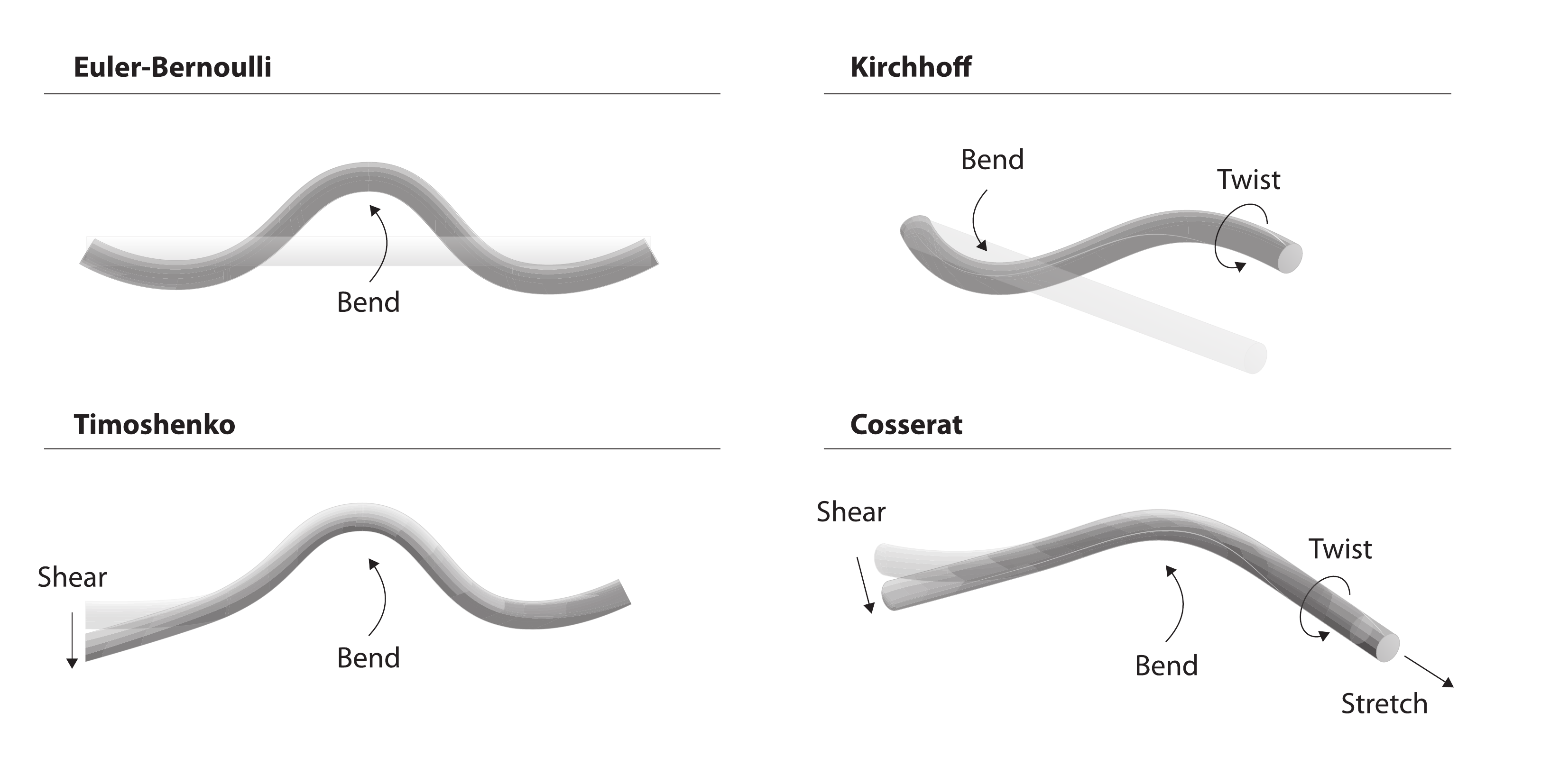} 
    \caption{Overview of rod theories. Euler-Bernoulli considers an elastic rod, supposed to be unstretchable and unshearable, that can only bend in one plane. Timoshenko extends the Euler-Bernoulli formulation by considering shear and bending. Kirchhoff introduces the concept of directors, modeling bending and torsion. From the directors' idea, Cosserat Rod Theory expands the Kirchhoff Rod Theory, including also linear deformations, such as shear and elongation.
    }
    \label{fig:rod_theories}
\end{figure}

An elastic rod is a \textit{quasi} \ac{1D} body whose length $L$ is much larger than the radius of its cross-section. Under this assumption, deformation is primarily described along the longitudinal axis, while cross-sectional deformations are neglected. Slender elastic structures are widespread in nature (e.g., hairs, muscle fibers, DNA strands, flagella), and provide an effective approximation for many continuum and soft robots.
Mathematically, the configuration of an elastic rod is represented by a time-varying curve, commonly referred to as the \textit{backbone}. The backbone is parameterized by the material curvilinear abscissa $s \in [0,L]$, where each point, denoted by $\bm{r}(s,t)$, is expressed with respect to an inertial frame $\{I\} = \{O_I; \bm{x}_I,\bm{y}_I,\bm{z}_I\}$. The points corresponding to $s=0$ and $s=L$ are referred to as the \textit{base} and \textit{tip} of the rod, respectively. Each cross-section is characterized by geometric and material properties, including the cross-sectional area $A(s)$, the second moment of area tensor $\bm{J} = \mathrm{diag}(J_x(s), J_y(s), J_z(s))$, the mass density $\rho(s)$, the Young's modulus $E(s)$, and the shear modulus $G(s)=E(s)/2(1+\nu)$, where $\nu$ denotes the Poisson ratio. In this work, Lie Algebra notation is frequently adopted, and the corresponding mathematical operators are reported in Appendix~\ref{sec:Appendix-B}.
Elastic rods may exhibit bending, torsion, stretching (elongation or compression), and shear. Different rod formulations capture different combinations of these deformation modes, leading to trade-offs between accuracy and computational complexity for simulation and control. \autoref{fig:rod_theories} summarizes the strain modes considered by the main rod theories employed in soft robotics, namely \textit{Cosserat--Reissner}, \textit{Kirchhoff--Clebsch--Love}, \textit{Euler--Bernoulli}, and \textit{Timoshenko--Ehrenfest}, which are reviewed in the following sections.

\subsection{Cosserat Rod Theory} \label{theoretical_background:cosserat}
The \ac{CRT} \cite{cosserat1896} describes all strain modes of the rod. This model is useful for describing \acp{CSM} that interact with cluttered environments, in which shear and elongation deformations play a fundamental role. Each cross-section $s$ is associated with a reference system $\{S_s\} = \{\bm{O}_s; \, \bm{x}_s, \bm{y}_s, \bm{z}_s\}$, whose axes are called \textit{directors}. The relative roto-translation between $\{S_s\}$ and $\{I\}$ is expressed by the homogeneous matrix $\bm{g}(s, t) \in SE(3)$ defined as
\begin{equation} \label{g_definition}
    \bm{g}(s, t) = \begin{bmatrix} \bm{R}(s, t) & \bm{r}(s, t) \\ \bm{0}^\top & 1 \end{bmatrix} \, ,
\end{equation}
where $\bm{R}(s, t) \in SO(3)$ is the rotation matrix that represents relative rotation. The main idea of the \ac{CRT} is to allow every cross-section $s$ to freely rotate and translate relatively.
\subsubsection{\textbf{Kinematics}}
Let the \textit{strain twist} $\bm{\xi}(s, t) \in \mathbb{R}^6$ be defined as
\begin{equation} \label{xi_definition}
    \bm{\xi}(s, t) = \left(\bm{g}^{-1}(s, t) \, \bm{g}'(s, t)\right)^{\vee} \, , \; \bm{\xi}(s, t) = \begin{bmatrix}
                            \bm{\kappa}(s, t) \\ \bm{\sigma}(s, t) 
                        \end{bmatrix} \, ,
\end{equation}
where $\bm{\kappa}(s, t) \in \mathbb{R}^3$ are the angular strain modes (bending and torsion), $\bm{\sigma}(s, t) \in \mathbb{R}^3$ are the linear strain modes (shear and elongation/compression), and $\left(\cdot\right)' = \frac{\partial}{\partial s}\left(\cdot\right)$ is the spatial partial derivative. 
Similarly, it is possible to define the \textit{velocity twist} $\bm{\eta}(s, t) \in \mathbb{R}^6$ as
\begin{equation} \label{eta_definition}
    \bm{\eta}(s, t) = \left(\bm{g}^{-1}(s, t) \, \dot{\bm{g}}(s, t)\right)^{\vee} \, , \; \bm{\eta}(s, t) = \begin{bmatrix}
                            \bm{\omega}(s, t) \\ \bm{v}(s, t) 
                        \end{bmatrix} \, ,
\end{equation}
where $\bm{\omega}(s, t) \textnormal{,} \, \bm{v}(s, t) \in \mathbb{R}^3$ are the angular and linear velocities of cross-section $s$, and $\dot{\left(\cdot\right)} = \frac{\partial}{\partial t}\left(\cdot\right)$ is the time partial derivative. The strain and velocity twists are both expressed in the local frame $\{S_s\}$ and represent the evolution of the rod over space and time.
Thanks to the mixed partial derivatives equality, it is possible to derive a kinematic relation between the strain and velocity twist, such as
\begin{equation} \label{cosserat_kinematics}
    \bm{\eta}(s) = \textnormal{Ad}_{\bm{g}^{-1}} \int_0^{s} \textnormal{Ad}_{\bm{g}} \, \dot{\bm{\xi}}(\zeta) \textnormal{d} \zeta \, .
\end{equation}
where $\zeta$ denotes an integration variable instead of the material abscissa $s$. \\
From a mathematical point of view, the main idea of the \ac{CRT} can be well condensed in the Configuration Space, defined as
\begin{equation} \label{cosserat_configuration_space}
    \mathcal{C} = SE(3) \times SE(3) \times \dots \times SE(3) \times \dots \, .
\end{equation}
Definition \eqref{cosserat_configuration_space} indicates the infinite \ac{DoFs} of an elastic rod. The Configuration Space results in a functional space of curves in $SE(3)$ \cite{boyer2020dynamics}. While the infinite-dimensional configuration space is a unique feature of soft robots, it also makes control particularly challenging.
\subsubsection{\textbf{Dynamics}}
To describe the \acp{EoM} of an elastic rod, it is necessary to define three distributed wrenches along the length of the rod:
\begin{itemize}
    \item The \textit{Internal Forces Wrench} $\bm{\mathcal{F}}_i = \begin{bmatrix} \bm{m}_i^{\top} & \bm{n}_i^{\top} \end{bmatrix}^{\top} \in \mathbb{R}^6$ expresses the internal load applied by the material, including elastic and damping effects. Assuming small strains, a linear viscoelastic constitutive model relates the strain field $\bm{\xi}$ and the internal forces wrench $\bm{\mathcal{F}}_i$.
    \begin{equation} \label{viscoelastic_law_cosserat}
        \bm{\mathcal{F}_i}(s) = \bm{\Sigma} \left(\bm{\xi} - \bm{\xi}^{*}\right) + \bm{\Upsilon} \dot{\bm{\xi}} \, ,
    \end{equation}
    where $\bm{\xi}^* \in \mathbb{R}^{6}$ is the stress-free strain twist, $\bm{\Sigma} = \textnormal{diag} \left(G J_x, E J_y, E J_z, E A, G A, G A \right) = \textnormal{diag} \left( \bm{\Sigma}_{\bm{\kappa}} , \, \bm{\Sigma}_{\bm{\sigma}} \right) \in \mathbb{R}^{6 \times 6}$ is the stiffness matrix and $\bm{\Upsilon} = \beta \, \textnormal{diag} \left(J_x, 3 J_y, 3 J_z, 3 A, A, A \right) \in \mathbb{R}^{6 \times 6}$ is the viscosity matrix.
    
    \item The \textit{External Forces Wrench} $\bm{\mathcal{F}}_e = \begin{bmatrix} \bm{m}_e^{\top} & \bm{n}_e^{\top} \end{bmatrix}^{\top} \in \mathbb{R}^6$ represents the distributed external load applied to the Rod. For instance, the effect of gravity can be computed as
    \begin{equation} \label{eq:gravity_terms_cosserat}
        \bm{\mathcal{F}}_e = \bm{\mathcal{M}} \left( \textnormal{Ad}_{\bm{g}}^{-1} \bm{\mathcal{G}} \right) \, ,
    \end{equation}
    where $\bm{\mathcal{G}} \in \mathbb{R}^6$ is the gravity acceleration twist w.r.t. the inertial frame and $\bm{\mathcal{M}} = \rho \, \textnormal{diag}\left(J_x, J_y, J_z, A, A, A\right)$ is the cross-sectional inertia matrix.

    \item The \textit{Actuation Forces Wrench} $\bm{\mathcal{F}}_a = \begin{bmatrix} \bm{m}_a^{\top} & \bm{n}_a^{\top} \end{bmatrix}^{\top} \in \mathbb{R}^6$ expresses the internal active forces exerted by the actuators. Refer to Sec. \ref{sec:actuation} for more details.
\end{itemize}
For clarity, the distributed wrenches ($\bm{\mathcal{F}}_i$, $\bm{\mathcal{F}}_a$, $\bm{\mathcal{F}}_e$) represent the forces and torques acting on a particular cross-section ($s = \bar{s}$), rather than force or torque densities distributed along the body's length.

After defining these quantities, it is possible to compute the Dynamics of an elastic rod using the Poincaré equations \cite{renda2018discrete}
\begin{equation} \label{compact_cosserat_dynamics}
    \bm{\mathcal{M}} \dot{\bm{\eta}} + \textnormal{ad}^*_{\bm{\eta}} \left(\bm{\mathcal{M}} \bm{\eta}\right) = \left(\bm{\mathcal{F}}_i - \bm{\mathcal{F}}_a \right)' + \textnormal{ad}^*_{\bm{\xi}} \left(\bm{\mathcal{F}}_i - \bm{\mathcal{F}}_a \right) + \bm{\mathcal{F}}_{e} \, .
\end{equation}
The \acp{EoM} \eqref{compact_cosserat_dynamics} is a set of \acp{PDE} of a Cosserat Rod in the local frame. It is common also to find the same set of \acp{PDE} in an explicit form and expressed in the inertial frame, such as \cite{gazzola2018forward}
\begin{equation} \label{expanded_cosserat_dynamics}
    \begin{split}
        \rho A \dot{\bm{v}}^I &= \left(\bm{n}_{i, a}^{I}\right)'  + \bm{n}_{e}^{I} \\
        \rho \bm{I} \dot{\bm{\omega}}^I + \bm{\omega}^I \wedge \left(\rho \bm{I} \bm{\omega}^I \right) &=  \left(\bm{m}_{i, a}^{I}\right)' + \bm{r}' \wedge \bm{n}_{i, a}^{I} + \bm{m}_{e}^{I}
    \end{split} \, ,
\end{equation}
where $\left(\cdot\right)^I$ denotes a variable expressed in the inertial frame, $\bm{I} = \bm{R} \bm{J} \bm{R}^{\top}$ is the second moment of the area expressed in the inertial frame, and $\bm{n}_{i, a} = \bm{n}_{i} - \bm{n}_{a}$ and $\bm{m}_{i, a} = \bm{m}_{i} - \bm{m}_{a}$ are the contributions of the internal and actuation forces.

\subsubsection{\textbf{Strain Parameterization}} \label{cosserat:strain_parameterization}
The Configuration Space \eqref{cosserat_configuration_space} and \acp{PDE} \eqref{compact_cosserat_dynamics} show the infinite \ac{DoFs} of an elastic rod. From a control and numerical implementation perspective, it is convenient to find a method to discretize the continuum body. \cite{renda2020geometric} and \cite{boyer2020dynamics} proposed a Strain Parameterization, also known as \ac{GVS} \cite{mathew2025reduced}, expressing the Kinematics and the Dynamics of the elastic rod in terms of strain twist $\bm{\xi}(s, t)$. Let us again consider the definition of the strain twist \eqref{xi_definition}, which is easily rewritten as $\bm{g}' = \bm{g} \, \hat{\bm{\xi}}$. Assuming that the initial condition $\bm{g}(0) = \bm{g}_0$ is known, the function $\bm{g}(s)$ can be univocally determined by the strain twist. The Configuration Space can be considered the \textit{Shape Space} $\mathbb{S}$ of the elastic rod, which is a functional space of the $s$-parameterized curves in $\mathbb{R}^6$, such as 
\begin{equation} \label{cosserat_configuration_space2}
    \mathcal{C} = \mathbb{S} = \{\bm{\xi}: s \in [0, L] \rightarrow \bm{\xi}(s) \in \mathbb{R}^6\} \, .
\end{equation}
This functional space can be generated by an infinite-dimension basis matrix $\bm{B}_{\bm{q}}(s)$, such as
\begin{equation} \label{gvs_discretization}
    \bm{\xi}(s, t) = \bm{B}_{\bm{q}}(s) \, \bm{q}(t) + \bm{\xi}^* \, , 
\end{equation}
where $\bm{B}_{\bm{q}} \in \mathbb{R}^{6 \times n}$ and $\bm{q} \in \mathbb{R}^n$ is a vector of generalized coordinates, with $n \rightarrow \infty$. The main idea of the discretization technique is to truncate the basis matrix to a finite number of $n$ columns, reducing the discretized Shape Space to $\mathcal{C} = \mathbb{R}^{n}$. This approach allows the user to choose the degree of approximation or neglect certain strain modes. To solve the \acp{PDE}, the proposed method uses the Magnus Expansion \cite{harier2006geometric}, resulting in a convenient Product of Exponentials, which is widely used in classical robotics \cite{murray}
\begin{equation} \label{renda_forward_kin}
    \bm{g}(s) = \bm{g}_0 \, \exp{\left(\hat{\bm{\Omega}}(s)\right)} \, ,
\end{equation}
where $\hat{\bm{\Omega}}(s) \in \mathfrak{se}(3)$ denotes the Magnus expansion of the strain twist.
Thanks to that, it is possible to rewrite the Differential Kinematics with a well-known form, such as
\begin{equation} \label{renda_diff_kin}
    \bm{\eta}(s, t) = \bm{J}(\bm{q}, s) \, \dot{\bm{q}} \, ,
\end{equation}
where $\bm{J}(\bm{q}, s) \in \mathbb{R}^{6 \times n}$ denotes the \textit{Soft Geometric Jacobian}.
Similarly, the Dynamics can be rewritten in a classical Lagrangian form
\begin{equation} \label{renda_dyn}
    \bm{M} \ddot{\bm{q}} + \bm{C} \dot{\bm{q}} + \bm{K} \bm{q} + \bm{D} \dot{\bm{q}} = \bm{B} \bm{\tau} + \bm{F}_{e} \, .
\end{equation}
The \acp{EoM} \eqref{renda_dyn} can be solved using a standard time solver, such as Runge-Kutta or Explicit Euler, as implemented in the SoRoSim simulator \cite{mathew2022sorosim}.

Recently, the Authors showed how to compute the analytical derivatives of the \ac{GVS} approach \cite{mathew2024analytical}, evolving SoRoSim as a differentiable simulator. In this work, the Authors also showed the benefits of the implicit time-integration schemes, such as the Implicit Euler and the Newmark-$\beta$ methods.

Finally, to fully describe soft robots interacting with the environment, \cite{xun2024cosserat} proposed an extension of \eqref{renda_dyn}, including frictional contacts with rigid and soft bodies.

\subsubsection{\textbf{Discrete Elastic Rod}}
\label{cosserat:discrete_elastic_rod}
Another method to discretize the continuum nature of elastic rods was introduced in the pioneering work of \cite{gazzola2018forward}. First, the Authors derived the \acp{EoM} \eqref{expanded_cosserat_dynamics}, including an elongation/compression ratio defined as $e = \textnormal{d} s / \textnormal{d} \bar{s}$, where $\bar{s}$ is the curvilinear abscissa in the rest configuration. This ratio is present because the length is parameterized using the curvilinear abscissa differently from \eqref{expanded_cosserat_dynamics}. In the presence of axial stretching, the stretching ratio $e$ scales the geometrical quantities
\begin{equation} \label{gazzola_rescaling}
    A = \frac{\bar{A}}{e}, \quad \bm{J} = \frac{\bar{\bm{J}}}{e^2}, \quad \bm{\Sigma} = \textnormal{diag}\left(\frac{1}{e^2}, \frac{1}{e} \right) \bar{\bm{\Sigma}}, \quad \bm{\kappa} = \frac{\bar{\bm{\kappa}}}{e}
\end{equation}
where the bar sign $\bar{\left(\cdot\right)}$ indicates the geometric quantities in the rest configuration. The \acp{EoM} can be finally written as
\begin{equation} \label{gazzola_dynamics}
    \begin{split}
        \rho A \dot{\bm{v}}^I &= \left(\frac{\bm{n}_{i, a}^{I}}{e}\right)' + e \, \bm{n}_{e}^{I} \\
        \rho \left(\frac{\bm{J}}{e}\right) \dot{\bm{\omega}} + \bm{\omega} \wedge \left(\rho \, \frac{\bm{J}}{e} \, \bm{\omega} \right) &= \left(\frac{\bm{m}_{i, a}}{e^3}\right)' + \frac{\bm{\kappa} \wedge \bm{m}_{i, a}}{e^3} \\
        &+ \bm{R}^{\top} \left(\frac{\bm{r}'}{e}\right) \wedge \bm{n}_{i, a} \\ 
        &+ \left(\rho \, \frac{\bm{J}}{e^2} \, \bm{\omega} \right) \dot{e} + e \, \bm{m}_{e}
    \end{split} \, .
\end{equation}
It is worth highlighting that \eqref{gazzola_dynamics} contains an additional contribution with respect to \eqref{compact_cosserat_dynamics}, which depends on the time derivative of stretching ratio $\dot{e}$. Furthermore, scaling the geometrical quantities partially relaxes the assumption of rigid cross-sections. To numerically resolve the \acp{EoM} \eqref{gazzola_dynamics}, they extended for a Cosserat rod the spatial discretization algorithm proposed in a previous study \cite{bergou2008discrete}, which discretizes the rod in a sequence of $N$ rigid segments connecting $N+1$ nodes. For each node, the \acp{EoM} consider the interactions with the other nodes and external forces. In addition, it is possible to associate kinematic and dynamic quantities to each node and segment to solve \eqref{gazzola_dynamics}. In the discrete domain, some quantities, such as curvature, must be expressed in an integrated form over the domain $\mathcal{D}$ \cite{gazzola2018forward}. This domain corresponds to the \textit{Voronoi} region $\mathcal{D}_i$ associated with the interior nodes $i \in [1, N-1]$,
\begin{equation}
    \mathcal{D}_i = (\ell_{i-1} + \ell_i)/2 \, ,
\end{equation}
where $\ell_i = |\bm{r}_{i + 1} - \bm{r}_{i}|$ is the length of the $i$-th segment. 
Then, the discrete curvature and bending stiffness matrix can be written as
\begin{equation} \label{discrete_curvature}
    \begin{split}
    \bar{\bm{\kappa}}_{i}&=\frac{\log\left(\bm{R}^{\top}_i \bm{R}_{i-1}\right)}{\bar{\mathcal{D}}_i} \\
        \bar{\bm{\mathcal{B}}}_i&= \frac{\bar{\bm{\Sigma}}_{\bm{\kappa}, i} \, \ell_{i} + \bar{\bm{\Sigma}}_{\bm{\kappa}, i - 1} \, \ell_{i-1}}{2\bar{\mathcal{D}}_i}
    \end{split},
\end{equation}
with $i \in [1, \, N - 1]$. Finally, the discretized \acp{EoM} of the rod can be rewritten in algorithmic form as follows:
\begin{equation} \label{gazzola_discrete_dynamics}
    \begin{split}
        m_i \dot{\bm{v}}^{I}_i &= \Delta^h \left( \frac{\bm{R}_j \left( \bar{\bm{\Sigma}}_{\bm{\sigma}, j} \bm{\sigma}_{j}\right)}{e_j} \right) + \bar{\bm{F}}_i \\
        \frac{ \bar{\bm{J}}_j}{e_j} \dot{\bm{\omega}}_{j} &= \Delta^h \left( \frac{\bar{\bm{\mathcal{B}}}_i \bar{\bm{\kappa}}_{i}}{\mathcal{E}_i^3} \right) + \mathcal{A}^h\left(\frac{\bm{\kappa}_{i} \wedge \bar{\bm{\mathcal{B}}}_i \bm{\kappa}_{i}}{\mathcal{E}_i^3} \mathcal{D}_i \right) + \\ 
        & + \left( \bm{R}^{\top}_j \, \bm{r}'_j \wedge \bar{\bm{\Sigma}}_{\bm{\sigma}, j} \bm{\sigma}_{j} \right)\bar{\ell}_j + \left( \bar{\bm{J}}_j \, \frac{\bm{\omega}_{j}}{e_j} \right) \wedge \bm{\omega}_{j} + \\ 
        & + \frac{\bar{\bm{J}}_j \bm{\omega}_{j}}{e_j^2} \, \dot{e}_j + \bm{C}_{j}
    \end{split} \, ,
\end{equation}
where $m_i$ is the point-wise mass associated with the node, $\mathcal{E}_i = \mathcal{D}_i / \bar{\mathcal{D}}_i$ is a domain dilation factor, $\Delta^h$ and $\mathcal{A}^h$ are the discrete difference and the averaging operator defined in \cite{gazzola2018forward}. In the first equation of \eqref{gazzola_discrete_dynamics}, $i \in [0, N]$, and $j \in [0, N-1]$. In contrast, in the second equation, $i$ ranges $[1, N-1]$, while $j$ still ranges from $[0, N-1]$. This distinction arises from the definition of discrete curvature and bending stiffness within the interior nodes set.

For the time integration of the discretized \acp{EoM} \eqref{gazzola_discrete_dynamics}, the authors proposed a \textit{Second-Order Position Verlet} time integrator, which exhibits a good balance between numerical accuracy and computational cost \cite{gazzola2018forward}. To ensure numerical stability, instead of enforcing rigorous Courant-Friedrichs-Levy stability conditions \cite{courant1967partial}, they proposed an empirical law to choose the integration time step as \mbox{$dt \approx 0.01 \frac{L}{N}$}. The discrete \acp{EoM} \eqref{gazzola_discrete_dynamics} were implemented in the PyElastica simulator \cite{naughton2021elastica}.

\subsection{Kirchhoff Rod Theory} \label{theoretical_background:kirchhoff}
    The \ac{KRT} \cite{Love1906} is a special case of \ac{CRT} that considers an elastic rod unstretchable and unshearable \cite{o2017modeling, gazzola2018forward, coleman1993dynamics}. It is particularly suitable for \acp{CSM} that bend around any axis and twist. Notably, it introduces the notion of a \textit{directed curve}, assigning a specific reference system to each cross-section. The \acp{EoM} of Kirchhoff rods can be computed by specializing the \acp{EoM} of Cosserat, that is \eqref{compact_cosserat_dynamics} or  \eqref{expanded_cosserat_dynamics}. In particular, the constraint of an unstretchable and unshearable rod can be written as
\begin{equation} \label{kirchhoff_constraint}
    \bm{\sigma} = \begin{bmatrix} 1 & 0 & 0 \end{bmatrix}^{\top} \, .
\end{equation}
The constraint \eqref{kirchhoff_constraint} enforces $e(t) \equiv 1$, which implies $\dot{e} = 0$ \cite{gazzola2018forward}. 
Incorporating this constraint into the \acp{EoM} leads to the following modified equations of motion:
\begin{equation} \label{kirchhoff_gazzola_dynamics}
    \begin{split}
        \rho A \dot{\bm{v}}^I &= \left(\bm{n}_{i, a}^{I}\right)' + \bm{n}_{e}^{I} \\
        \rho \bm{J} \dot{\bm{\omega}} + \bm{\omega} \wedge \left(\rho \bm{J} \bm{\omega} \right) &= \left(\bm{m}_{i, a}\right)' + \bm{\kappa} \wedge \bm{m}_{i, a} \\ 
        &+ \left(\bm{R}^{\top} \bm{r}'\right) \wedge \bm{n}_{i, a} + \bm{m}_{e}
    \end{split} \, .
\end{equation}

The linear internal force $\bm{n}_{i, a}$ serves as a Lagrangian multiplier and it is a virtual internal force that constrains the rod from stretching or shearing. As stated in \cite{gazzola2018forward}, inserting the constraint of an unshearable and unstretchable rod can increase the computational time.

In the case of Strain Parameterization, it is sufficient to apply the constraint \eqref{kirchhoff_constraint} in the strain twist
\begin{equation} \label{eq:kirchhoff_gvs}
    \bm{\xi}(s, t) = \begin{bmatrix} \bm{\kappa}^{\top}(s, t) & \begin{pmatrix} 1 & 0 & 0 \end{pmatrix}^{\top} \end{bmatrix}^{\top} \, .   
\end{equation}
Therefore, the basis function matrix only generates the angular strain mode vector $\bm{\kappa}(s, t)$. Unlike the previous approach, \eqref{eq:kirchhoff_gvs} does not apply as a constraint, and the \acp{EoM} can be solved without considering any Lagrangian multipliers.
    
\subsection{Euler-Bernoulli Rod Theory} \label{theoretical_background:euler_bernoulli}
The \ac{EBRT} \cite{timoshenko1983history} is one of the simplest rod theories in which the rod can only bend around one axis. It can be considered the 2D case of the \ac{KRT} without the twist. Here, the assumption is that the slope angle of the rod is equal to the tangent angle of the backbone curve. Below, we report linear and nonlinear versions.

\subsubsection{\textbf{Linear Euler-Bernoulli}}
\label{euler_bernoulli:linear}
Let us consider the rod in the $x$-$y$ plane. Unlike the previous theories, the backbone curve is described by the displacement $w(x, t) \in \mathbb{R}$ from the $x$-axis. Recalling the assumption of the EBRT, the slope angle $\alpha(x, t) \in \mathbb{R}$ of the beam can be written as
\begin{equation} \label{euler_bernoulli:slope_angle}
    \alpha(x, t) = \frac{\partial w}{\partial x} \, .
\end{equation}
From the minimization of the strain energy, the \ac{EoM} of the rod can be computed as
\begin{equation} \label{EB_dynamics}
    E J_z \frac{\partial^{4} w(x, t)}{\partial x^4} + \rho A \frac{\partial^{2} w(x, t)}{\partial t^2} = m_{e, z}(x, t) \, ,
\end{equation}
where $m_{e, z}(x, t)$ is the distributed external moments around the $z$-axis.
The \ac{EoM} \eqref{EB_dynamics} is linear, assuming that the constitutive law of the bending moment $m_{i, z}(x, t)$ is
\begin{equation} \label{EB_costitutive_law}
    m_{i, z} = E J_z \frac{\partial^2 w}{\partial x^2} \, ,
\end{equation}
where $m_{i, z} \in \mathbb{R}$ is the $z$-component of the internal moment $\bm{m}_i$.
For control purposes, the linearity of \eqref{EB_dynamics} facilitates the use of efficient controllers from the classic control theory \cite{doroudchi2018decentralized}.

\subsubsection{\textbf{Nonlinear Euler-Bernoulli}}
\label{euler_bernoulli:nonlinear}
In the nonlinear EBRT (or Euler's Elastica), the rod length is parameterized by the arclength $s$. Recalling the \textit{Fundamental Theorem of the local theory of Curves} in the 2D case, every regular curve can be determined by the curvature \cite{carmo_differential_geometry}. From this concept, \cite{della2019control} reformulated Elastica in a classical robotic formulation. Let $\kappa_z(s)$ be the $z$-component of the curvature twist $\bm{\kappa}(s)$. The Cartesian pose of every cross-section $s$ is uniquely determined by the curvature, i.e.
\begin{equation} \label{pol_curv_cartesian}
    \begin{split}
        x(s) &= L \, \int_{0}^{s} \cos\left(\alpha(\zeta)\right) \, \textnormal{d} \zeta \, , \\
        y(s) &= L \, \int_{0}^{s} \sin\left(\alpha(\zeta)\right) \, \textnormal{d} \zeta \, ,\\
        \alpha(s) &= \int_{0}^{s} \kappa_{z}(\zeta) \, \textnormal{d} \zeta \, .
    \end{split}
\end{equation}
Furthermore, from \eqref{pol_curv_cartesian}, it is possible to derive the EoM following the Lagrangian approach. 
Similarly to the Strain Parameterization, they propose to consider the curvature $\kappa_z(s)$ as an infinite sum of monomials in $s$, i.e.
\begin{equation} \label{polynomial_curvauture}
    \kappa_z(s) = \sum_{i = 0}^{n-1} \theta_i \left(\frac{s}{L}\right)^i \quad \textnormal{with} \; n \rightarrow \infty  \, .
\end{equation}
The geometrical meaning of the polynomial curvature is to constrain the shape of the backbone curve to be a Generalized Cornu Spiral, i.e., a curve with a polynomial curvature \cite{della2019control}. It is worth highlighting that the \eqref{polynomial_curvauture} is equivalent to write \eqref{gvs_discretization} with a polynomial basis function, related only to the curvature $\kappa_z$ \cite{caradonna2024model}.
Defining $\bm{q} = \begin{bmatrix} \theta_0 & \theta_1 & \cdots & \theta_{n-1} \end{bmatrix}^{\top} \in \mathbb{R}^{n}$ as joint variables, it is possible to write the \acp{EoM} as
\begin{equation}
    \bm{M}\ddot{\bm{q}} + \bm{C}\dot{\bm{q}} + \bm{G} + \bm{K} \bm{q} + \bm{D} \dot{\bm{q}} = \bm{A}\left(\bm{q}\right) \bm{\tau} \, ,
\end{equation}
where $\bm{A}(\bm{q}) \in \mathbb{R}^{n \times n_a}$ is transposed orientation jacobian for $n_a$ actuators and $\bm{\tau} \in \mathbb{R}^{n_a}$ are the pure actuators' torque.

 \subsection{Timoshenko Rod Theory} \label{theoretical_background:timoshenko}
    The Timoshenko-Ehrenfest rod theory \cite{Timoshenko1951} extends the \ac{EBRT}, relaxing the unshearability constraint. Therefore, the equality between the tangent angle and the cross-sections' angle is no longer valid (i.e., $\alpha(x) \neq {\partial w}/{\partial x}$). For this reason, the \acp{EoM} of a Timoshenko rod considers the angle $\alpha(x, t)$ as an independent variable. Its dynamics can be written as
\begin{equation} \label{eq:timoshenko_dynamic}
    \begin{split}
        \rho A \left(\frac{\partial^{2} w}{\partial t^{2}}\right) &= m_{e, z}(x,t) + \frac{\partial}{\partial x} n_{i, y}(x,t) \\
        \rho J_z \left(\frac{\partial^{2}\alpha}{\partial t^{2}}\right) &= \frac{\partial}{\partial x} m_{i, z}(x,t) + n_{i, y}(x,t).
    \end{split}
\end{equation}
The standard formulation assumes a linear constitutive law with bending moment $m_{i, z}$ and shear force $n_{i, y}$ defined as
\begin{equation} \label{timoshenko_constitutive_law}
    m_{i, z} = E J_z \, \frac{\partial \alpha}{\partial x} \qquad 
    n_{i, y} = \gamma \, G A \left(\frac{\partial w}{\partial x}-\alpha\right),
\end{equation}
where $\gamma$ is the Timoshenko shear coefficient, which depends on the cross-section geometry. We can see that the shear force contribution is proportional to the difference between the tangent angle $\partial w / \partial x$ and the angle of cross-section $\alpha$.

\section{Modeling Actuation}
\label{sec:actuation}
\begin{figure}[t]
    \centering
    \includegraphics[width=0.9\columnwidth]{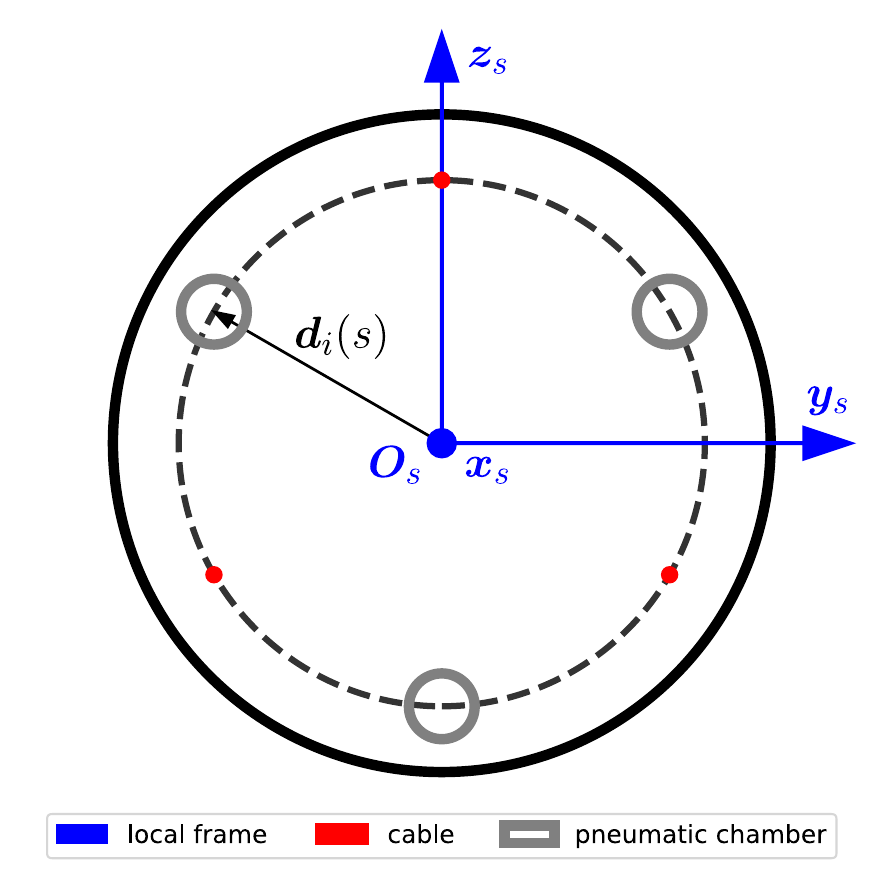} 
    \caption{Cross section of a soft robot with equally spaced actuators (e.g., pneumatic chambers, cables). The vector $\bm{d}_i(s)$ is the position of the $i$-th actuator in the local frame $\{S_s\}$.}
    \label{fig:cross_section}
\end{figure}

The distinction between physical and robotic models lies in the inclusion of actuation. In soft robotics, actuation sources greatly differ from those used in rigid robots. In particular, the most used technologies are cables and fluidic chambers, which guarantee distributed active loads $\bm{\mathcal{F}}_a(s)$.

\subsection{The Actuation Matrix}
Consider a Cosserat rod with $n_a$ actuators. Let be $\bm{d}_{i}(s) \in \mathbb{R}^{3}$ the distance from the center of the cross-section $s$ and the $i$-th actuator (\autoref{fig:cross_section}). Each actuator applies an internal active wrench that depends on the actuator routing. We can express the internal active wrench $\bm{\mathcal{F}}_a^{(i)}$ applied by the actuator $i$ as
\begin{equation} \label{actuation_model}
    \bm{\mathcal{F}}_a^{(i)} = \begin{bmatrix} \tilde{\bm{d}}_i(s) \bm{t}_i(s) \\ \bm{t}_i(s) \end{bmatrix} \, \tau_i \, ,
\end{equation}
where $\bm{t}_i(s)$ is the unit vector tangent to the actuator path and $\tau_i$ is the magnitude of the actuator. For example, $\tau_i$ can be
\begin{equation}
    \tau_i = \begin{cases}
                T & \textnormal{if cable-driven actuation} \\
                p A_{\textnormal{in}} & \textnormal{if fluidic actuation}
            \end{cases} \, ,
\end{equation}
where $T$ is the cable tension, $p$ is the pneumatic pressure and $A_{{\textnormal{in}}}$ is the internal cross-section area of the fluidic chamber. The expression of the tangent unit vector is derived as
\begin{equation}
    \bm{t}_{i}(s) = 
    \frac{\left[\bm{g}^{-1} \left(\bm{g} \bm{d}_i\right)'\right]_3}{\lVert{\left[\bm{g}^{-1} \left(\bm{g} \bm{d}_i\right)'\right]_3}\rVert} = 
    \frac{\left[\hat{\bm{\xi}}(s) \bm{d}_i(s) + \bm{d}'_{i}(s)\right]_3}{\lVert{\left[\hat{\bm{\xi}}(s) \bm{d}_i(s) + \bm{d}'_{i}(s)\right]_3}\rVert} \, ,
\end{equation}
where $\bm{d}_i(s)$ is expressed in homogeneous coordinates and the operator $\left[\cdot\right]_3$ extracts the first three rows of a homogeneous vector.
The resultant of the contributions of $n_a$ actuators is 
\begin{equation} \label{resultant_active_forces}
        \bm{\mathcal{F}}_a(s) = \sum_{i = 1}^{n_a} \bm{\mathcal{F}}_a^{(i)}(s) = \bm{B}_{\bm{\tau}}(s) \, \bm{\tau} \, ,
\end{equation}
where $\bm{\tau} \in \mathbb{R}^{n_a}$ in the internal active wrench and $\bm{B}_{\bm{\tau}} \in \mathbb{R}^{6 \times n_a}$ is the actuation matrix
\begin{equation} \label{actuation_matrix}
    \bm{B}_{\bm{\tau}}(s) = 
    \begin{bmatrix} 
    \begin{pmatrix} 
    \tilde{\bm{d}}_i(s) \bm{t}_i(s) \\
    \bm{t}_i(s) 
    \end{pmatrix}_{i=1}^{n_a}
    \end{bmatrix}.
\end{equation}
The actuation matrix $\bm{B}_{\bm{\tau}}(s)$ is crucial for designing control algorithms because it contains valuable information about the strain modes induced by the actuators. From a control perspective, this information is related to the system’s \textit{reachability}.

\subsection{Actuators-Strain Mapping}
From the definition of actuation matrix \eqref{actuation_matrix}, it is possible to map the actuation magnitude vector to the distributed active load $\bm{\mathcal{F}}_{a}$. \cite{renda2020geometric} propose a statics-based method to relate the actuators with excited strain modes. In particular, let be the statics of a Cosserat Rod as
\begin{equation} \label{compact_cosserat_statics}
    \left(\bm{\mathcal{F}}_i - \bm{\mathcal{F}}_a\right)' + \textnormal{ad}^{*}\left(\bm{\mathcal{F}}_i - \bm{\mathcal{F}}_a\right) = \bm{0} \, ,
\end{equation}
assuming no external forces and the strain twist is discretized with the Strain Parameterization approach, i.e. \eqref{gvs_discretization}. By invoking the D'Alembert Principle, it is possible to derive
\begin{equation}
    \int_{0}^{L} \bm{B}^{\top}_{\bm{q}} \bm{\mathcal{F}}_i \, \textnormal{d} s = \int_{0}^{L} \bm{B}^{\top}_{\bm{q}} \bm{\mathcal{F}}_a \, \textnormal{d} s \, .
\end{equation}
Finally, substituting the constitutive relation \eqref{viscoelastic_law_cosserat} and the definition of actuation forces \eqref{resultant_active_forces}, we derive
\begin{equation}
    \left(\int_{0}^{L} \bm{B}^{\top}_{\bm{q}} \bm{\Sigma} \bm{B}_{\bm{q}} \, \textnormal{d} s \right) \bm{q} = \left(\int_{0}^{L} \bm{B}^{\top}_{\bm{q}} \bm{B}_{\bm{\tau}} \, \textnormal{d} s\right) \bm{\tau} \, .
\end{equation}
From this equality, \cite{renda2024dynamics} propose a \textit{trivial} form of the static equation, choosing the implicit parametrization 
\begin{equation} \label{trivial_gvs}
    \bm{\xi}(s) - \bm{\xi}^*(s) = \bm{\Sigma}^{-1}(s) \bm{B}_{\bm{\tau}}(\bm{\xi}, s) \, \bm{q} \, .
\end{equation}
Consequentially, the static equation degenerates in a trivial form, such as
\begin{equation} \label{trivial_static}
    \begin{split}
        \left(\int_{0}^{L} \bm{B}^{\top}_{\bm{q}} \bm{B}_{\bm{\tau}} \, \textnormal{d} s \right) \bm{q} &= \left(\int_{0}^{L} \bm{B}^{\top}_{\bm{q}} \bm{B}_{\bm{\tau}} \, \textnormal{d} s\right) \bm{\tau} \, , \\ \bm{q} &= \bm{\tau} \, .
    \end{split}
\end{equation}
The functional basis derived from the implicit parametrization \eqref{trivial_gvs} uses the information of actuation routing contained in $\bm{B}_{\bm{\tau}}$, to provide the minimum set of functional bases. 
In particular, these functional bases describe the excited strain modes and their optimal static representation. Furthermore, $\bm{B}_{\bm{q}}$ in \eqref{actuation_matrix} also considers the geometrical and the material information of the robot, contained in $\bm{\Sigma}$. 
Finally, \eqref{trivial_gvs} provides a useful starting point for constructing the optimal functional basis matrix $\bm{B}_{\bm{q}}$, avoiding redundant or inefficient shape functions while preserving accuracy and computational efficiency.

\section{Rod Models for Continuum and Soft Robots}
\label{sec:models}

\begin{figure}[t]
    \centering
    \includegraphics[width=0.9\linewidth]{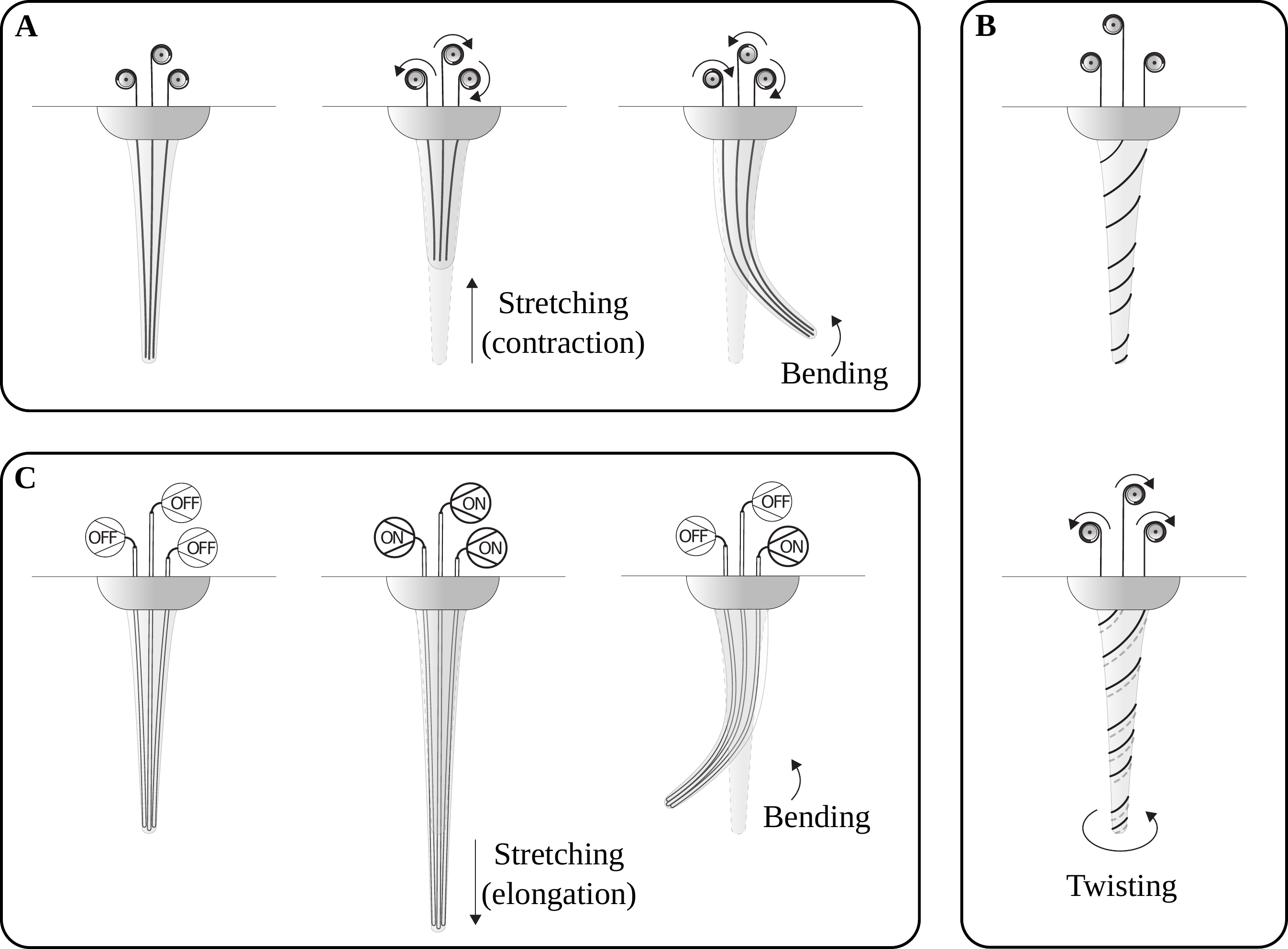}
    \caption{Principal deformations exhibited by common continuum and soft manipulators. \textbf{A-B}) Cable-driven manipulators stretch (contraction), bend, and twist, based on the actuators' path along the centerline and the cross-section. \textbf{C}) Similarly, pneumatic manipulators mostly stretch (elongate) and bend.}
\label{fig:robot_sketch}
\end{figure}

We categorize rod models based on the deformation modes they represent, independently of the underlying actuation technology. From a control perspective, the specific actuation method (e.g., tendons, pneumatic chambers) is secondary to the type of strain it produces, such as elongation or contraction, torsion, or bending. The admissible deformation modes also implicitly determine the suitable rod theory. For example, a robot limited to planar bending is consistent with the Euler-Bernoulli framework, while more complex deformation behaviors might require Cosserat theory. While several studies formally employ the Cosserat rod formulation, in practice, they restrict the model to a limited subset of strain modes, effectively reducing it to simpler formulations. By explicitly classifying models according to the combination of strain modes they allow, we highlight such modeling choices. Out of the fifteen possible combinations of deformation modes, only nine classes emerged. We summarize the deformation classes in \autoref{tab:unified_model_table} to facilitate the literature comprehension, where we arbitrarily report representative articles to distill each class and highlight the versatility of rod models. An advanced Scopus query aided the initial selection of the articles (see Appendix~\ref{sec:Appendix-A}). We also discussed articles beyond the query.

\subsection{Bend}
Bending is a prevalent deformation in continuum and soft robots due to the natural propensity of slender structures to bend (\autoref{fig:robot_sketch}). Various rod models were employed to predict the bending motion for different actuation mechanisms, stiffness modulation strategies, or dynamic interactions with the environment. To clarify the different approaches, we distinguish between planar (2D) bending and spatial (3D) bending.

\textbf{Planar bending (2D)}: Planar bending assumes that deformation occurs in a single plane, which significantly simplifies the mathematical formulation. Among the approaches, the \ac{EBRT} is particularly suitable for modeling simple soft robots that bend due to its simplicity. Indeed, \ac{EBRT} modeled various robotic architectures, achieving sub-millimeter tip position accuracy in multi-backbone continuum robots \cite{he2013analytic} and enabling shape modeling in \acp{PCM} via its non-linear formulation \cite{li2018shape}. 
The Timoshenko rod theory, which is suitable for planar deformations (\autoref{fig:rod_theories}), was used to model planar bending in cable-driven soft robots with complex geometries, such as triangular notched structures \cite{wenlong2013mechanics}.
Pneumatically actuated soft robots also benefited from rod models. \ac{EBRT} captured the relationship between internal pressure and bending deformation \cite{de2016constitutive}, later extended to handle large deflections and non-linear strains \cite{janizadeh2023steady, wu2022design}. \cite{sadati2019reduced} introduced a \ac{ROM} and a discretized formulation of the STIFF-FLOP robot using \ac{EBRT}, demonstrating high accuracy, numerical robustness, and near real-time performances. \ac{EBRT} has also been applied for modeling planar \ac{PCM} with two independent flexible panels \cite{li2023analytic}, integrating geometric constraints due to the moving platform. 
Beyond kinematics, \cite{gao2016mechanical} developed a Cosserat rod model for continuum manipulators, incorporating external forces, tendon forces and friction, with experimental validation. 
Dynamic planar bending has also been addressed through variational approaches. In particular, \cite{tariverdi2020dynamic} proposed a Lie group variational integration framework experimentally validated on magnetically actuated \acp{CSM}, achieving sub-millimeter tip accuracy. 

Planar bending models were also applied to investigate locomotion strategies. \cite{zhou2015flexing} modeled peristaltic locomotion with an Euler–Bernoulli rod, and \cite{cicconofri2015study} used a planar Cosserat rod with adjustable curvature to reproduce snake-like swimming. As a further extension, \cite{daily2017dynamical} developed a dynamic Euler–Bernoulli model of a caterpillar-inspired robot with \ac{SMA} actuators, linking undulation mechanics to ground contact forces. Similarly, \cite{xiang2019study} applied an Euler–Bernoulli formulation to predict rolling and crawling in a magnetically actuated soft robot. These works prove the versatility of rod models beyond manipulation.

\textbf{Spatial bending (3D)}:
Spatial bending occurs when deformation is not confined to a single plane but involves curvature around multiple axes, eventually accompanied by torsion. These cases typically require Cosserat or Timoshenko formulations to accurately capture the complete range of strain modes.
\ac{CRT}, despite its higher complexity, has extensively contributed to bending modeling in cable-driven robots \cite{jones2009three_dim}.
An early example of \ac{CRT}-based bending modeling is provided by \cite{giorelli2012two}, who modeled the kinematics of a cable-driven soft arm with non-uniform cross-sections and derived its inverse model using the Jacobian matrix.

In pneumatic actuators, \ac{CRT} was coupled with nonlinear constitutive laws (e.g., Mooney–Rivlin hyperelasticity) to improve accuracy \cite{Roshanfar2023Hyperelastic}. Similarly, \cite{namdar2022static} computed the static shape of a semicylindrical soft fiber-reinforced pneumatic bending actuator under external force constraints, using a Neo-Hookean model to derive the pressure-angle relationship in free motion \cite{ogden2013}.
\ac{CRT} was also applied to soft robots embedding smart actuators \cite{Pawlowski2018modeling, Pawlowski2018dynam_model, doroudchi2020dynamic}.
Beyond kinematic descriptions, dynamic models mainly based on \ac{CRT}, provide more suitability for dynamic tasks and allow the integration of actuation into its formulation \cite{Ma2021dynamicsModel}. 
Further refinements in \ac{CRT}-based cable-driven robot models included friction effects arising from interactions between cables and the robot structure \cite{wang2023general}. 
\ac{CRT} was also leveraged to model tasks involving interaction with external objects. \cite{jalali2021dynamic} introduced a \ac{CRT}-based framework to analyze the coupled dynamics of tendon-driven cooperative continuum robots interacting with flexible objects.
Likewise, \cite{dehghani2014dynamics} proposed a dynamic Cosserat rod model for a tendon-driven continuum robotic finger, where the grasping forces are modeled as external forces acting on the backbone.
These studies highlight the effectiveness of \ac{CRT} in dynamic interaction tasks, including tip force estimation in cable-driven catheters, as demonstrated by \cite{hooshiar2021analytical}, who used a Bézier-based shape interpolation and an optimization approach.
Another important aspect of dynamic interactions concerns workspace estimation, namely the determination of the spatial region that the soft robot’s end-effector is able to reach given its actuation and mechanical constraints. \cite{amehri2021workspace} developed a quasi-static \ac{CRT} model with a continuation algorithm to map workspace boundaries in tendon-driven manipulators, addressing self-collision, elastic instabilities, and actuator saturation. The study found that complex \ac{CSM} configurations could lead to isolated boundaries and voids, which are essential to consider for accurate workspace predictions.
Beyond kinematic and dynamic predictions, stiffness modulation also plays an important role in predicting soft robots behavior \cite{xiao2023kinematics} as well as mitigating undesired effects such as buckling, which can hinder precise movements \cite{molaei2022cable}. Notable advancements involve variable stiffness continuum manipulators, which, leveraging \ac{CRT}-based modeling can increase stiffness up to ten times, resulting in improved control precision \cite{zhao2020sim}. Other techniques include the addition of layer jamming sheaths to control stiffness both along the axial and transverse directions \cite{fan2022novel}. 
In parallel, research on a simple rod-shaped silicone soft robot, modeled using \ac{CRT}, investigated stiffness properties by applying horizontal pulling forces at the free ends of the rod \cite{grube2022simulation}. Two material models were examined: a linear viscoelastic model and a nonlinear Saint Venant-Kirchhoff material model. The damping mechanism was analyzed, revealing that linear damping can be assumed for small to medium oscillations. However, non-linear effects become relevant when oscillations reach higher amplitudes, enhancing the model's accuracy.
\cite{Nalkenani2021modeling} presented a \ac{CRT} model of a soft bending actuator with an additional fabric layer to modify the actuator stiffness, modeled as boundary condition of the Cosserat model. This adjustment improved bending behavior while preventing longitudinal stretching.

\ac{FEM} was widely used to augment and validate rod-based models, particularly for pneumatic actuators. For instance, \cite{de2017development} proposed a five-parameter constitutive relation for a pneumatic soft arm, with parameters identified via \ac{FEM} simulations under different loading conditions. Conversely, \cite{mbakop2021inverse} integrated \ac{PH} curves with an \ac{EBRT} framework to reconstruct actuator shapes from virtual control points, establishing a link between control inputs and geometry. Both approaches, as well as more recent formulations \cite{li2023analytic}, were validated through \ac{FEM} and experimental results, confirming their accuracy. In parallel, \cite{rone2013continuum} developed a high-fidelity lumped-parameter dynamic model for cable-driven continuum robots, based on the principle of virtual power. It captured curvature variations while accounting for inertial, actuation, friction, elastic, and gravitational effects.

\subsection{All Deformations}
\label{sec:all_deformations}
As soft robotics advanced, the research community explored the use of continuum and soft robots in unstructured environments. In this context, the \ac{CRT} emerged as a fundamental framework for modeling their deformation. Indeed, interaction with irregularly shaped objects makes shear effects significant, while advances in actuation technologies require the modeling of stretching and torsional deformations. A recent contribution by \cite{Tuummers2023Cosserat} further consolidates \ac{CRT} as a standard framework for continuum robots. The authors rigorously derived equivalent Newtonian and Lagrangian formulations for multi-segment tendon-driven robots, providing practical guidelines for numerical implementation.

As discussed in Sec.~\ref{sec:theoretical_background}, the \ac{CRT} describes the motion of an elastic rod with an infinite \ac{DoFs}. Discretization techniques play a key role in numerically solving the governing equations of Cosserat rods. Several methods were developed to enhance numerical robustness and computational efficiency. For example, \cite{renda2018discrete} introduced the \ac{PCS} method to discretize a continuous Cosserat rod into segments with constant strain. This method extends the \ac{PCC} approach \cite{webster2010design} by incorporating shear and torsion. In parallel, \cite{till2019real} proposed a real-time discretization technique that transforms the \acp{PDE} of a Cosserat rod into a \ac{BVP} by discretizing time derivatives, significantly improving computational performance.

Incorporating external forces such as buoyancy, drag, and added mass is essential for accurately modeling the interactions between soft robots and their surroundings. In this context, \cite{renda20123d, renda2014dynamic} analyzed the static and dynamic behavior of a silicone arm in a dense medium, capturing its coupled tendon conditions and external loads. The model was further developed in a unified Cosserat rod framework applied to underwater robots \cite{renda2018unified}. By accounting for nonlinearities and hydrodynamic effects, this approach enabled the evaluation of soft robot designs under submerged conditions.

For pneumatic actuators, \cite{renda2017screw} extended the \ac{PCS} method by integrating Screw Theory, allowing flexible routing of fluidic and tendon actuators along arbitrary paths. In \acp{PCM}, \cite{armanini2021discrete} adapted the \ac{PCS} formulation to closed-chain soft robotic systems, facilitating the modeling and design optimization of structures like the Fin-Ray finger.

Material nonlinearities, such as hysteresis, can significantly affect the performance of soft robots. \cite{mishra2022fractional} incorporated nonlinear material effects by proposing a fractional-order Bouc-Wen model within a Cosserat rod. The model demonstrated superior accuracy in capturing the large deformations of pneumatic soft actuators compared to traditional approaches.

Extensive validation of \ac{CRT} models is essential for ensuring their predictive accuracy. Cable-driven soft robots were widely studied and experimentally validated. For example, \cite{renda20123d} validated a static Cosserat model of a tendon-driven soft arm using \ac{FEM} simulations and experimental comparisons. The dynamic extension of this model \cite{renda2014dynamic} was experimentally verified by replicating characteristic octopus movements. Later, \cite{renda2018unified} demonstrated the effectiveness of their unified dynamic framework using an underwater robotic system. Further validation of discretization-based approaches was performed in \cite{renda2018discrete} and \cite{till2019real}, confirming the numerical robustness of their methods. Conversely, \cite{butler2019continuum} investigated deflection and stiffness in robots subject to constant axial tendon displacement and environmental loads, comparing parallel and converging actuator paths. They validated the Cosserat rod model on a \ac{CSM} subject to tip loads in the $0 - 0.05$ kg range, providing actuations to reach bending angles up to $60^\circ$. The simulation results aligned well with the experimental data. However, as the authors noted, limitations emerged for larger curvatures due to the increased role of friction, which was not included in the model.
For pneumatic soft robots, \cite{trivedi2007geometrically, trivedi2008geometrically} developed geometrically exact \ac{CRT} models and validated them on fiber-reinforced pneumatic manipulators, achieving significantly higher accuracy than constant curvature models. The screw-theoretic \ac{PCS} formulation \cite{renda2017screw} was validated on the STIFF-FLOP arm, while \cite{Roshanfar2021towards} tested their pneumatic Cosserat rod model experimentally, reporting an $8.7\%L$ tip position error.
Also, \ac{FEM} was extensively used to compare and validate \ac{CRT} models. Indeed, \cite{grazioso2018analytic} derived analytical solutions for quasi-static Cosserat rods and verified them against \ac{FEM} simulations under external loads. Similarly, \cite{berthold2021preliminary} conducted a comparative study between \ac{FEM} and \ac{CRT} models for pneumatic actuators, proposing a parameter identification process. Further refinement was made by \cite{wiese2022towards}, who obtained a static Cosserat model from detailed \ac{FEM} simulations, validating it experimentally across various loading conditions.
For \acp{PCM}, \cite{till2017elastic} investigated elastic stability using Cosserat rod models and experimentally validated their stability conditions. In parallel, \cite{black2017parallel} introduced a kinetostatic \ac{CRT} framework for \acp{PCM}, achieving high accuracy in force sensing and manipulability analysis. Then, \cite{lilge2022kinetostatic} extended this approach, developing a comprehensive CRT-based framework for tendon-driven \acp{PCM}, experimentally achieving a median pose accuracy of $3.4\%L$.

Cosserat rod models found widespread applications in bioinspired robotics and biomechanical studies. \cite{boyer2017locomotion} proposed a unified framework for modeling the locomotion of bioinspired robots with soft appendages. They analyzed flapping flight and passive swimming, comparing two different modeling approaches based on Newton-Euler dynamics. Similarly, \cite{zhang2019modeling} implemented Cosserat rod simulations in PyElastica \cite{naughton2021elastica}, assembling heterogeneous structures to replicate musculoskeletal architectures (e.g., snakes, bird wings). Their work demonstrated the effectiveness of \ac{CRT} in bioengineering design, simulating complex musculoskeletal interactions in biological and robotic applications. Then, \cite{boyer2022statics} explored the intersection of \ac{CRT} and optimal control theory, revealing fundamental insights into the singularity of dynamic continuum robot simulations. By reformulating the problem as a minimization task, their approach provided a novel perspective on the statics and dynamics of continuum robots, validated through simulations of bioinspired continuum swimmers.

Finally, to find optimal bases for the \ac{GVS} approach, \cite{alkayas2024soft} introduced a data-driven reduction method based on \ac{POD}. Their approach begins by creating a high-fidelity \ac{GVS} digital twin from experimental data, which is then simulated to generate a comprehensive dataset of strains. By applying \ac{SVD} to this data and truncating the least significant singular values, the model's dimensionality is reduced while preserving its accuracy. This process yields an optimal linear basis of the form $\bm{B}_{bm{q}}\left(s\right)$. A subsequent study extended the search to the nonlinear case (i.e., $\bm{B}_{\bm{q}}\left(\bm{q}, s\right)$), employing autoencoders to identify a more expressive and compact representation \cite{alkayas2025structure}.


\subsection{Bend \& Twist}
%
%
%
%
%
%
%
%
%
%
Bending and twisting are fundamental deformation modes that enable \acp{CSM} to achieve complex \ac{3D} shapes and dexterous movements. Modeling these modes is particularly important in applications where interaction with the environment is not limited to planar motions but requires full spatial maneuverability. In many cases, incorporating twisting alongside bending provides the best trade-off between accuracy and computational efficiency. This allows continuum robots to follow arbitrary curves in space, which is essential for tasks such as wrapping around an object, executing dexterous maneuvers, or navigating constrained environments. Biomedical applications especially benefit from bending-twisting modes, as in \acp{CTR} for \ac{MIS} \cite{rucker2010model,rucker2010geometrically,till2020dynamic} and \acp{ETR} for multi-arm sheaths \cite{wang2019steering,wang2020eccentric,wang2021eccentric}. Safety-critical domains also utilize bending-twisting compliance analysis to predict deformations under external loads \cite{smoljkic2014compliance}.

Twisting deformations are usually induced by particular cable or tendon routing. Tendon-driven continuum robots employ non-straight tendon paths that wrap around the backbone to generate both bending and twisting. Early work by \textit{Rucker et al.} \cite{rucker2011statics} established Cosserat rod formulations where curved tendon paths produce distributed loads along the backbone. Similarly, \cite{dehghani2013modeling} developed a \ac{CSM} with three tendons spiraling around a nitinol backbone, extending their previous untwisted setup \cite{dehghani2011mod_cont}. They further refined this approach for real-time control using a stable moment-based algorithm \cite{dehghani2014compact}. Twisting was also demonstrated in robotic tails \cite{rone2014continuum}, octopus-inspired manipulators \cite{xu2017modelling,niu2019closed}, tendon-actuated extensible robots \cite{chikhaoui2019comparison}, and multi-section tendon robots \cite{janabi2021cosserat}. In fluidic actuation, torsional effects are less common but can be induced through asymmetric or fiber-reinforced chambers. For example, \textit{Uppalapati et al.} \cite{uppalapati2018parameter} modeled a \ac{CSM} with two asymmetric pneumatic actuators capable of bending and rotating, which were later extended with an additional rotating actuator \cite{uppalapati2021design}. Other approaches include models based on Euler-Bernoulli for micro-tube pneumatics \cite{ji2019rapid} and generalized PneuNet structures, combining bending and twisting predictions validated against \ac{FEM} \cite{gu2021analytical}.

Cosserat rods were extensively applied to tendon-driven systems, while Euler-Bernoulli or energy-based methods are often preferred for pneumatic networks due to their simpler kinematics. Examples include asymmetric pneumatic actuators capable of torsion \cite{uppalapati2018parameter,uppalapati2021design}, micro-tube pneumatics modeled with multi-segment Euler-Bernoulli rods \cite{ji2019rapid}, and PneuNet actuators modeled analytically with energy minimization \cite{gu2021analytical}. These models capture how chamber geometry, reinforcement, and pressurization induce coupled bending and twisting modes. However, the majority of bending-twisting continuum robots are tendon-driven, concentric-tube, or eccentric-tube systems. These architectures require a geometrically exact description of distributed loads and torsional effects, which is naturally expressed through the \ac{CRT}.

Concerning \acp{CTR}, \cite{renda2021sliding} adapted the piecewise variable-strain \cite{mathew2025reduced} to simulate multi-section \acp{CTR} by including the tubes' sliding motion. Rotation motions of the tubes are included as generalized coordinates rather than boundary kinematic conditions. Notably, this procedure led to a minimal set of closed-form algebraic equations solvable for both the shape variables, the actuation forces, and the torques. The resulting equations also facilitate control, design optimization, and stability assessment. Unlike beam simplifications, Cosserat rods account for coupling between bending and torsion and handle spatially varying loads from tendon paths, external wrenches, and contact. This explains why most bending-twisting works in the literature adopt the Cosserat rod formalism \cite{rucker2011statics,dehghani2013modeling,dehghani2014compact,rone2014continuum,xu2017modelling,niu2019closed,chikhaoui2019comparison,janabi2021cosserat,wang2019steering,wang2020eccentric,wang2021eccentric}.

\textbf{Buckling}: Buckling instability is another important nonlinear phenomenon to consider. When modeling thermally activated \ac{TCP} muscles, \cite{wu2020modeling} employed \ac{KRT}, since the small-strain assumptions and direct link to coil load-twist relations simplify the actuation model. Similarly, Kirchhoff rods were used in micrometer-scale parallel continuum robots \cite{mauze2021micrometer} and bacteria-inspired flagellated swimmers exploiting buckling instabilities for propulsion \cite{forghani2021control}. Conversely, Cosserat rods were applied to study global buckling in soft robotic arms \cite{sipos2020longest}, where large deflections and intrinsic curvature control are essential. Thus, Kirchhoff rods often appear in actuation-centric or microscale studies where analytical tractability is crucial, while Cosserat rods dominate in large-deformation, shape-control analyses.

\textbf{Workspace analysis}: Finally, workspace analysis requires accurate models of bending and twisting under actuation and loading. Tendon-driven Cosserat models enable forward kinematics and Jacobian-based control for workspace prediction \cite{dehghani2014compact}. For pneumatic robots, static workspace matching was validated experimentally with asymmetric actuators modeled by Cosserat rods \cite{uppalapati2018parameter,uppalapati2021design}. PneuNet and micro-tube actuator models \cite{ji2019rapid,gu2021analytical} allow for workspace design and optimization by linking actuator geometry and pressure inputs to achievable curves. In biomedical contexts, concentric and eccentric tube models predict reachable poses in confined environments \cite{rucker2010model,wang2021eccentric}.


\begin{table*}[t]
\caption{Overview of rod-based models of continuum and soft robots.}
\label{tab:unified_model_table}
\small\sf\centering
\resizebox{0.95\textwidth}{!}{  
\begin{tabular}{@{}lllll@{}}
\toprule
\textbf{Deformation Class} &
\textbf{Subclass} &
\textbf{Ref.}  & 
\textbf{Contribution}  &
\textbf{Rod Model} \\ 
\midrule
\multirow{9}{*}{\textbf{Bend}} & \multirow{3}{*}{Cable} & \cite{jones2009three_dim} & Static model for real-time shape estimation of \acp{CSM}. & \ac{CRT}\\
 &  & \cite{dehghani2014dynamics} &  Dynamic model of a continuum finger for planar grasping. & \ac{CRT}\\
 &  & \cite{gao2016mechanical}  & Model of notched continuum manipulator for \ac{MIS}. & \ac{CRT}\\
\cmidrule{2-5}
 & \multirow{2}{*}{Pneumatic} & \cite{de2017development}  & Derivation of constitutive law of a planar bending actuator. &  \ac{EBRT}\\
 &  & \cite{sadati2019reduced}  & \ac{ROM} and rod models with absolute and relative states validated on STIFF-FLOP. & \ac{EBRT}\\
\cmidrule{2-5}
 & \multirow{2}{*}{Smart} & \cite{daily2017dynamical}  & Dynamic model inspired by caterpillar motion to analyse ondulation mechanics. & \ac{EBRT}\\
 & & \cite{Pawlowski2018modeling}  & Modeling the coupling between \ac{CSM} body and thermal actuation. & \ac{CRT}\\
\cmidrule{2-5}
& Parallel & \cite{li2018shape} & Shape modeling of a \ac{PCM} made of soft panels.  & \ac{EBRT}\\
\cmidrule{2-5}
& Tubular & \cite{pattanshetti2019kinematic}  & Estimation of stiffness variation on a monolithic tube for \ac{MIS}. & \ac{CRT}\\
\midrule
\multirow{9}{*}{\textbf{All Deformations}} & \multirow{3}{*}{Cable} & \cite{rucker2011statics} & Derived exact models for the forward kinematics, statics, and dynamics of \acp{CSM} with general tendon routing. &\multirow{9}{*}{\ac{CRT}} \\
& & \cite{renda2014dynamic} & Studied the dynamic interaction of an octopus-inspired \ac{CSM} with a dense medium. & \\
& & \cite{butler2019continuum} & Predicted deflection and stiffness in \acp{CSM} subject to constant actuation and external loads. & \\
\cmidrule{2-4}
  &  Pneumatic & \cite{trivedi2008geometrically}  & Model validation of \ac{CSM} with fiber-reinforced actuators varying base orientation. & \\
\cmidrule{2-4}
  & \multirow{3}{*}{Parallel} & \cite{till2017elastic} & Studied and experimentally validated the elastic stability of \acp{PCM}. & \\
  &  & \cite{armanini2021discrete} & Modeled closed-chain soft \acp{PCM} with validation on a soft-rigid finger. & \\
  &  &  \cite{lilge2022kinetostatic} & Studied reachability, singularities, manipulability, and compliance of \acp{PCM}. & \\
\cmidrule{2-4}
 & \multirow{2}{*}{Bioinspired} & \cite{zhang2019modeling} & Simulates musculoskeletal systems assembling heterogeneous active/passive rods. & \\
  &  &  \cite{boyer2022statics} & Studied the link between \ac{CRT} and optimal control simulating locomotors and swimmers. & \\
\midrule
\multirow{7}{*}{\textbf{Bend \& Twist}} & Cable & \cite{janabi2021cosserat}  & Tutorial on the dynamic modeling of tendon-driven continuum robots that bend and twist. & \ac{CRT}\\
\cmidrule{2-5}
& Pneumatic & \cite{uppalapati2021design}  & Design and modeling of a \ac{CSM} with parallel asymmetric fiber-reinforced elastomers.  & \ac{CRT}\\
\cmidrule{2-5}
& \multirow{3}{*}{Tubular} & \cite{till2020dynamic} & Dynamic model of \acp{CTR} validated for tissue grasping and snapping.  & \ac{CRT}\\
  &  & \cite{wang2021eccentric}   & Modeling of a superelastic \ac{ETR} sheath for biomedical procedures. &  \ac{CRT}\\
  & & \cite{renda2021sliding} & Adapted the piecewise variable strain to \acp{CTR} including the tubes' sliding motion. & \ac{CRT}\\
\cmidrule{2-5}
  & Smart     & \cite{wu2020modeling}     & \acp{TCP} muscles incorporating coil load and twist into actuation model. &  \ac{KRT}\\
\cmidrule{2-5}
  &  Buckling        &  \cite{sipos2020longest}    & Study relationship between buckling and $\bm{\kappa}$ on simulated cantilever rods. &  \ac{CRT}\\ 
\midrule
%
\multirow{9}{*}{\textbf{Bend \& Stretch}} & \multirow{2}{*}{Cable} & \cite{xiao2023kinematics}  & Addressed stiffness regulation for a soft arm with a sliding backbone, achieving +57.7\% stiffness.  & \ac{CRT} \\
  &  &  \cite{chen2022model}  & Augments the \ac{PCC} with a rod model to quickly solve the \ac{CSM} deformation under $\bm{\mathcal{F}}_e$. &  \ac{EBRT} \\
\cmidrule{2-5}
  & \multirow{5}{*}{Pneumatic} & \cite{sadati2017control}  &  Utilizes the Ritz-Galerkin method to reduce the continuous state space of rod models. & \ac{CRT} \\
  & & \cite{gilbert2019validation} & Experimentally validated of a \ac{CSM} model on bending motions generated by square pulse actuations. & \ac{CRT}\\
  & & \cite{Garbulinski2022bending} & Characterized bending, natural frequency, and damping of a cantilevered extensile artificial muscle. & \ac{EBRT}\\
  &  &  \cite{alessi2023ablation} &  Ablation study of model addressing manufacturing uncertainties in a 3D-printed \ac{CSM}.  & \ac{CRT}\\
  &  &  \cite{dou2023design}  &  Design and modeling of a hybrid \ac{CSM} incorporating an antagonistic compliant mechanism. &  \ac{EBRT}\\
\cmidrule{2-5}
  & Smart &  \cite{goldberg2019planar}  & Simulated locomotion of a caterpillar-inspired soft robot actuated by \acp{SMA}. &  \ac{CRT}\\
  \cmidrule{2-5}
  & Rod-driven & \cite{wu2022design} & Design, statics modeling, and numerical workspace analysis of an extensible rod-driven \ac{PCM}. & \ac{CRT}\\
\midrule
\multirow{6}{*}{\textbf{Bend \& Twist \& Stretch}} & \multirow{4}{*}{Cable} & \cite{renda2012general} & 
Simulated the coupled tendon drive for multi-section \acp{CSM} highlighting the role of twist. & \multirow{4}{*}{\ac{CRT}}\\
  &  & \cite{wang2016three} & Compared the Kelvin viscoelastic material model vs a purely elastic model on a real siliconic arm. & \\
  &  & \cite{adagolodjo2021coupling} & Combined \ac{FEM} to model the robot body with \ac{CRT} to model actuation. & \\
  & & \cite{Roshanfar2023Hyperelastic} & Investigated pneumatic stiffness regulation on a cable-driven \ac{CSM} using the hyperelastic material model. &  \\
\cmidrule{2-5}
  & Tubular & \cite{ryu2018application} & Reconstructed the shape of a continuum endoscope given external forces exerted on the distal end $\bm{\mathcal{F}}_e(L)$. & \ac{CRT}\\
\cmidrule{2-5}
  & Parallel & \cite{ghafoori2020modeling} & Modeled a six-link \ac{PCM} starting from an experimental validation of a single elastic rod. & \ac{CRT} \\
\midrule
\multirow{2}{*}{\textbf{Bend \& Stretch \& Shear}} & Tubular & \cite{chen2021variable} & Experimentally validates a model for multi-backbone continuum robot. &  \multirow{2}{*}{\ac{CRT}}\\
  & Pneumatic & \cite{bartholdt2021parameter} & Models a soft pneumatic actuator combining rods and \acp{ANN} & \\
\midrule
\multirow{2}{*}{\textbf{Bend \& Twist \& Shear}} & --- & \cite{grazioso2019geometrically} & Proposed a novel \ac{FEM}-inspired discretization technique and the SimSoft simulator. & \multirow{2}{*}{\ac{CRT}}\\ 
  & Tubular & \cite{bentley2023safer} & Proposed a force model and associated cost metric for safer and clinically relevant for motion planning. & \\ 
\midrule
\multirow{1}{*}{\textbf{Stretch \& Shear}} & Bioinspired  & \cite{hemingway2021continuous} & Studied the peristaltic locomotion of the earthworm \textit{Lumbricus terrestris} to model soft robots. & \ac{CRT}\\
\midrule
\multirow{1}{*}{\textbf{Stretch}} & Bioinspired & \cite{Hyousse2019distributed} & Studied the peristaltic locomotion of a caterpillar-inspired soft robot. & \ac{CRT} \\
\bottomrule
\end{tabular}
}
\end{table*}

\subsection{Bend \& Stretch}
Simultaneous bending and stretching behaviors are frequent in \acp{CSM} actuated by cables or pneumatics (\autoref{fig:robot_sketch}). Notably, this category excludes twist and shear effects for several reasons: the actuators are linear and apply forces along the centerline, not generating torsional or shear stresses; interactions with the environment are either neglected or primarily induce axial and bending loads. As a result, the deformation behavior can be described in terms of bending and axial stretching. We summarize several \textit{Bend \& Stretch} works in \autoref{tab:unified_model_table}. Herein, we focus the discussion on manufacturing variability, stiffness regulation, and fiber-reinforced actuators. 

\textbf{Manufacturing variabilities}: Manufacturing variability, as well as material degradation, are critical concerns in soft robotic systems, as they can significantly affect actuator behavior, model accuracy, and control performance. Variations in geometry, material stiffness, or the positioning of actuators during fabrication can cause substantial differences between the modeled behavior and the robot’s actual motion. While several experimentally validated rod models were developed for pneumatic soft arms \cite{sadati2017geometry, sadati2017mechanics, sadati2017control, gilbert2019validation, Wang2022DynamicsMod, Lamping2023ANovel} and fingers \cite{sachin2022analytical, flores2022soft}, relatively few have focused specifically on the effects of manufacturing irregularities. Below, we discuss two works using Cosserat rod models to address manufacturing uncertainties in pneumatic soft robots. 
\textit{Eugster et al.} \cite{eugster2022soft} described the kinematics of a soft arm using a nonlinear pressure-dependent constitutive law, the principle of virtual work, and modeled the actuator with strain energy functions. Manufacturing imperfections are considered by scaling the relations of extensional and bending stiffness and by a constant shift of the actuator position $\bm{d}_i(s)$. Assuming the pressure dilates the chambers, the model also considers cross-section deformations with a pressure-dependent diameter $D(p)$. The model is validated experimentally for static stretching and bending motions, in vertical and horizontal mountings, comparing constant or pressure-dependent chamber radius, with or without chamber repositioning, and a linear or Ogden material law. Similarly, \textit{Alessi et al.} \cite{alessi2023ablation} presented a dynamic model for a 3D-printed \ac{CSM} with actuators that exhibited different stretching when subject to equal actuation. They captured this effect by tuning the pressure-strain relation $\epsilon(p)$ for each actuator introducing \textit{strain gains} $\gamma^{(i)}$ that tune the pressure-strain relation $\epsilon(p)$ for each actuator

\begin{equation}
\epsilon(p^{(i)}) = \gamma^{(i)} \frac{p^{(i)} A_{\textnormal{in}}^{(i)}}{EA},
\end{equation}

where $A_{{\textnormal{in}}}$ is the interior actuator area. An ablation study conducted on dynamic motions investigated the contribution of different model components, uncovering that neglecting manufacturing uncertainties can indeed degrade performance up to $5\%L$.
In practice, manufacturing irregularities can affect every soft robotic system but remain unexplored outside the \textit{Bend \& Stretch} class. We attribute this to the fact that, for linear actuators, it is easier to quantify the expected motion and thus identify model discrepancies.

\textbf{Stiffness regulation}: Stiffness modeling and regulation are central to advancing the adaptability and precision of soft robotic systems. However, in many designs, limitations such as single-mode actuation, the absence of antagonistic mechanisms, or the lack of a structural backbone can restrict the ability to modulate stiffness effectively. To address these challenges, \textit{Sun et al.} \cite{Sun2018hybrid} presented a novel design of a hybrid continuum robot whose actuators combine pneumatic muscles with embedded elastic rods. The robot can regulate its stiffness through a locking mechanism, switching between large-scale movement enabled by pneumatics and fine positioning enabled by push-pull of the rods. Stiffness tests revealed that the robot increases its structural stiffness by 65\% during fine positioning (locked mode) and reduces the repetitive positioning error by 62\%. To model this behavior,  they improved a static \ac{KRT} calculating the elastic deformation through the minimal total potential energy principle in an optimal control framework.
Similarly, \cite{xiao2023kinematics} addressed stiffness regulation for a tendon-driven soft \ac{CTR} with a nitinol tube backbone, which can slide inside the soft body for pose or stiffness regulation. A Cosserat rod model was validated in several scenarios by varying the joint-space tendon inputs and the task-space external contact force, achieving an average tip position error below 1\% of the total length. Simulation studies demonstrate that inserting the nitinol backbone can enhance the kinematic workspace and increase stiffness by 57\%. Two manipulation case studies demonstrated the potential application of hybrid actuation for stiffness regulation. 
Another compliant rod-driven mechanism enabling stiffness regulation was introduced in a soft arm with pneumatic actuation \cite{dou2023design}. While an Euler-Bernoulli model was adapted to demonstrate the benefit of the mechanism in experiments of way-point tracking, the investigation of the stiffening behavior remains future work. 

\textbf{Fiber-reinforced actuators}: Fiber-reinforced soft pneumatic actuators present significant modeling challenges due to nonlinear deformation under pressure. Several works leveraged variations of rod theories to more accurately capture their complex bending, elongation, and cross-sectional behaviors. \textit{Sadati et al.} \cite{sadati2017geometry, sadati2017mechanics, sadati2017control} performed comprehensive investigations of several modeling approaches---experimentally validated on the well-known STIFF-FLOP arm---addressing the challenge of cross-section deformation, captured analytically through geometry deformation and \ac{CRT}. This approach outperformed constant curvature approaches by $13\%$ and variable curvature by $7\%$. 
More recently, \cite{Berthold2022investigation} introduced a Cosserat rod model to deepen the understanding of the pressure-induced deformation of a fiber-reinforced soft pneumatic actuator with three \ac{DoFs}. Specifically, they considered the compression effects of lateral pneumatic chamber walls by converting the inner chamber pressure into an equivalent force. Their model, validated against \ac{FEM} simulations, obtained a mean tip position error up to $11\%$. Conversely, \cite{Hanza2023mechanics} accounted for the effect of radial pressure by proposing an inhomogeneous Cosserat rod model with a nonlinear strain-force relationship. The proposed method outperformed the standard \ac{CRT} by about $15\%$. Meanwhile, \cite{LiuShengkai2023Modeling} explored the effects of self-gravity and external loads on the configuration of an actuator with a semicircular cross-section through an \ac{EBRT}. 







\subsection{Bend \& Twist \& Stretch}
A few works explored the combination of bending, twisting, and stretching, especially in cable-driven systems. 
\cite{renda2012general} presented a general geometrically exact static \ac{CRT} for cable-driven \acp{CSM}. The work enabled the simulation of coupled tendon drives for multiple sections through various design parameters and highlighted the role of torsional deformation. Although the study was conducted only in simulation, it provided a solid foundation for future full dynamic analyses (Sec.~\ref{sec:all_deformations}).
Similarly, \cite{wang2016three} modeled a cable-driven conic \ac{CSM} made of silicone by combining a geometrically exact dynamic Cosserat rod with the Kelvin viscoelastic material model. Experimental validation with real-world data included tests with the arm fixed in a cantilever configuration under gravity, without actuation ($\bm{\tau} = \bm{0}$) or executing 2D and 3D motions through different cable actuation combinations ($\bm{\tau} \neq \bm{0}$). The authors found that the Kelvin model matched the experimental data better than a purely elastic model.
Also, \cite{adagolodjo2021coupling} described the dynamics of cable-driven soft robots by combining \ac{FEM} for modeling the robot structure with a discrete Cosserat rod representation for the actuation system. The differential equations were integrated using an implicit backward Euler time-stepping scheme to ensure numerical stability, while both direct and inverse simulations demonstrated the suitability of the approach for robots actuated by cables or rods.
Recently, \textit{Roshanfar et al.} \cite{Roshanfar2023Hyperelastic} investigated stiffness regulation in a cable-driven \ac{CSM} equipped with a central linear pneumatic chamber. The authors captured both robot motion and pressure-induced stiffening behavior by augmenting the classical Cosserat rod formulation with a hyperelastic material model. 

In conclusion, while the \textit{Bend \& Twist \& Stretch} class comprises fewer works than the four main deformation classes, it has consistently contributed to the modeling of cable-driven systems. This class may be regarded as a particular case of the \textit{All Deformation} category, in which shear strains are assumed negligible and therefore omitted. Additional works belonging to this class are reported in \autoref{tab:unified_model_table}.

\subsection{Bend \& Stretch \& Shear}
The \textit{Bend \& Stretch \& Shear} deformation class, similar to the previous class, emerges by neglecting torsion.
\textit{Chen et al.} \cite{chen2021variable} proposed a variable curvature static-kinematic Cosserat rod model for multi-backbone continuum robots, incorporating multi-backbone structural constraints. The model is validated on a robot with two multi-backbone continuum segments, each with four nitinol rods. A push-pull actuation bends the continuum robot and transmits shear forces. The experiments evaluate the tip positioning accuracy for circular trajectories and the shape discrepancy when the robot is subject to an end-point load in two configurations. They also implement an inverse kinematics in a real-time open-loop controller for tracking a rectangular path, with and without tip load. The proposed model outperforms the \ac{PCC}. Note that the torsion-free assumption did not hold in all experiments. Indeed, when out-of-plane external forces were applied to cause torsion, the tip position error raised to 19\% of the length.
Differently, \cite{bartholdt2021parameter} developed a hybrid model combining the classical \ac{CRT} with a data-driven stiffness estimation. Rather than solving numerically nonlinear constitutive equations, the model relies on a work point-dependent linear stress-strain relationship for computational efficiency. Experimental validation is carried out by identifying the stretching and bending stiffness of a soft pneumatic actuator and learning the actuation to stiffness maps using \acp{ANN}.


\subsection{Bend \& Twist \& Shear}
Another special case is when robots bend, twist, and shear, without stretching. \textit{Grazioso et al.} \cite{grazioso2019geometrically} proposed a \ac{FEM}-inspired spatial discretization technique where the rod is divided into several nodes, defining a helicoidal shape function for the interpolation. Thanks to the specific choice of the shape function, they formulated the Cosserat rod dynamics with Lie Groups. The proposed method was implemented in a new simulator called SimSOFT and validated on the pure bending of a cantilever soft arm and the pure in-plane rotation of a soft arm with varying external conditions. In addition, they experimentally validated the coupling between bending, torsion, and shear on the Princeton arm benchmark, achieving excellent accuracy and computational efficiency.
Shifting to clinical applications, \textit{Bentley et al.} \cite{bentley2023safer} derived a tissue and needle force model with a Cosserat string formulation. It describes normal and frictional forces along the shaft as a function of the planned needle path, the friction model with its parameters, and the piercing force. The proposed force model and associated cost metric are safer and more clinically relevant for motion planning. They fit and validate the model through physical needle robot experiments in a gel phantom. The force model defines a cost function for motion planning, and was evaluated against a path-length cost function in random environments, reducing the peak force by 62\%. 


\subsection{Axial Deformations}
Axial deformations (stretching and shearing) are common in peristaltic locomotion, where waves of contraction, extension, and shear generate propulsion along the robot body.

\textbf{Stretch \& Shear}: The peristaltic locomotion of earthworms was studied with a continuous model of compressible and incompressible slender bodies represented as Cosserat rods \cite{hemingway2021continuous}. Incompressibility is enforced as an internal constraint using Green and Naghdi’s theory of a directed rod. Two linearly elastic isotropic material models are assumed, with material parameters identified experimentally for small deformations. Motion effects from actuators or muscle contraction are modeled as external compressive loads using a doublet function for an assigned centerline force and a uniform pressure for a pair of assigned director forces. The method is showcased in simulation on a soft robotic device.

\textbf{Stretch}: In a bioinspired robotics application, \cite{Hyousse2019distributed} modeled a caterpillar soft robot using a network of linear and torsional springs connected by massless rods represented as planar discrete elastic rods \cite{goldberg2019planar}.



\begin{figure}[t]
    \centering
    \includegraphics[width=\linewidth]{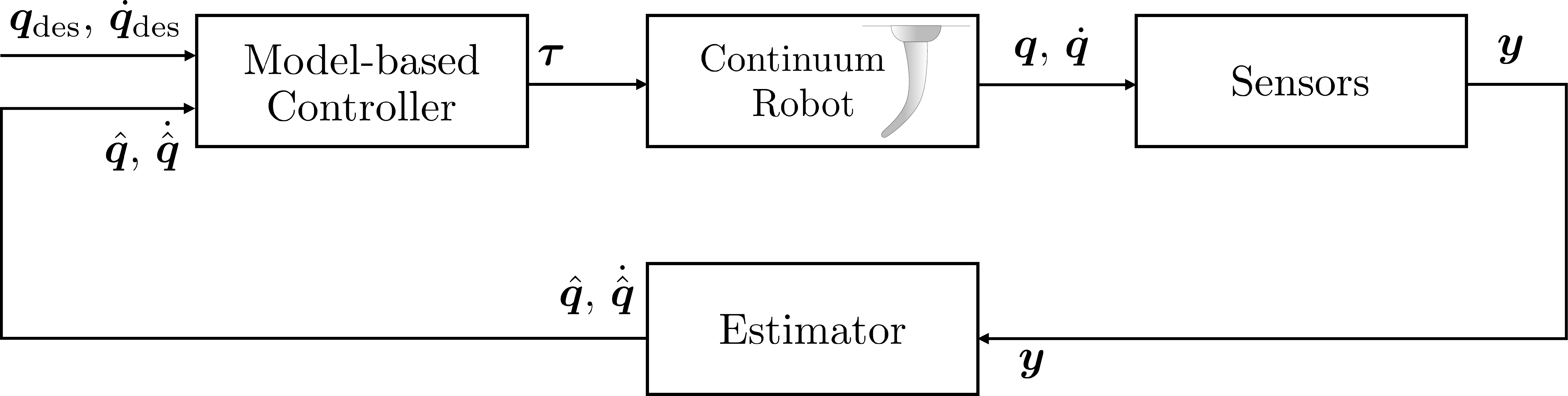} 
    \caption{General block diagram of the \ac{MB} control architecture.}
    \label{fig:mb_scheme}
\end{figure}

\acresetall 
\section{Model-based Controllers for Rod Models} 
\label{sec:model_based_controllers}
Controlling continuum and soft robots can be challenging due to material nonlinearities (e.g., hysteresis), the infinite-dimensional configuration space, and under-actuation. These properties make the implementation of well-known controllers for rigid manipulators difficult or impossible.
The control literature for rigid robots primarily focuses on controlling the end-effector in terms of pose, velocity, or force. For continuum robots, the natural equivalent task is tip control. To enable continuum and soft robots to achieve capabilities beyond those of rigid robots, researchers have proposed two additional control tasks:
\begin{itemize}
\item \ac{MP} control, which regulates the pose of a set of cross-sections,
\item Shape control, which regulates the robot backbone, fully utilizing the potential of continuum structures.
\end{itemize}

The \ac{MB} control strategy involves leveraging prior knowledge of the robot to guide it toward achieving a specific task. 
The motivation behind developing \ac{MB} controllers is the possibility of mathematically guaranteeing the stability and performance of the controlled system. 
However, implementing controllers in physical prototypes can be challenging, often requiring an extensive identification procedure. In this context, the spatial discretization technique is crucial to balance accuracy and computational efficiency. Therefore, the model selection and the discretization technique are integral parts of the control design process.

\textbf{Formalization}: 
To formalize the control problem, let us consider $\bm{q} \in \mathbb{R}^{n}$ a generic configuration vector, whatever discretization algorithm is used. To formalize the control problem, let us consider $\bm{q} \in \mathbb{R}^{n}$ a generic configuration vector, admitting any discretization algorithm. According to \cite{della2023model}, the shape control problem can be particularized in two sub-problems: (i) regulation and (ii) \ac{TT}. In the \ac{TT} case, the goal of shape control is to find an input $\bm{\tau}(t)$ such that
\begin{equation} \label{shape_control_formulation}
    \lim_{t \to \infty} \begin{bmatrix} \bm{q}(t) \\ \dot{\bm{q}}(t) \end{bmatrix} = \begin{bmatrix} \bm{q}_{\textnormal{des}}(t) \\ \dot{\bm{q}}_{\textnormal{des}}(t) \end{bmatrix} \, ,
\end{equation}
where $\bm{q}_{\textnormal{des}}, \, \dot{\bm{q}}_{\textnormal{des}} \in \mathbb{R}^{n}$ are the desired configuration and its time derivative, respectively. The condition \eqref{shape_control_formulation} can be particularized in the regulation case, imposing $\dot{\bm{q}}_{\textnormal{des}} = \bm{0}$.
Similarly to rigid robots, it is possible to formalize the tip control problem, rewriting \eqref{shape_control_formulation} in the task space, such as
\begin{equation} \label{tip_control_formulation}
     \begin{split}
      & \lim_{t \to \infty} \bm{g}(L, t) = \bm{g}_{\textnormal{des}}(L, t) \\
      & \lim_{t \to \infty} \bm{\eta}(L, t) = \bm{\eta}_{\textnormal{des}}(L, t)      
    \end{split} \, ,   
\end{equation}
where $\bm{g}_{\textnormal{des}}(L, t) \in SE(3)$ and $\bm{\eta}_{\textnormal{des}}(L, t) \in \mathbb{R}^6$ are the desired pose and velocity twist of the tip. Similarly, the tip pose regulation condition can be obtained, particularizing \eqref{tip_control_formulation} with $\bm{\eta}_{\textnormal{des}}(L, t) = \bm{0}$.
Lastly, the \ac{MP} control problem can be formalized by extending \eqref{tip_control_formulation} in a set of desired poses and velocity twists of the cross sections.

To address these control problems, the \ac{MB} control framework can be represented through the general scheme shown in \autoref{fig:mb_scheme}. The block ``Sensors" refers to the sensors mounted on the robotic platform, mapping $\bm{q}, \, \dot{\bm{q}}$ in the measurements $\bm{y} \in \mathbb{R}^p$. The ``Estimator" block implements the Shape Estimation/Reconstruction algorithms, providing an estimation of $\bm{q}$, $\dot{\bm{q}}$, defined as $\hat{\bm{q}}$, $\dot{\hat{\bm{q}}} \in \mathbb{R}^n$. Finally, the MB controller block computes the input $\bm{\tau}$, processing the error between the estimated configuration and the desired one.
Following this general scheme, many works on \ac{MB} control occurred, thanks to the development of accurate and computationally efficient rod models. Below, we review the contributions, while \autoref{tab:mb-controllers-table} reports the key points.

\subsection{Collocated Form}
To address the underactuated nature of continuum robots, a recent method proposed by \cite{pustina2024input} reformulates the dynamic model \eqref{renda_dyn} in a collocated form. This approach decouples the actuated variables from the passive ones via a change of coordinates, $\bm{\theta} = \bm{h}\left(\bm{q}\right)$. The resulting state vector $\bm{\theta} \in \mathbb{R}^{n}$ is partitioned into actuated coordinates, $\bm{\theta}_{a} \in \mathbb{R}^{n_a}$, and passive coordinates, $\bm{\theta}_{u} \in \mathbb{R}^{n - n_a}$.

A valid change of coordinates $\bm{h}\left(\bm{q}\right)$ can be expressed as
\begin{equation}
    \bm{h}\left(\bm{q}\right) = \begin{bmatrix}
                                    \bm{g}\left(\bm{q}\right) \\ \bm{0}_{n - n_a}
                                \end{bmatrix} + \begin{bmatrix} \bm{0}_{n_a \times n_a} & \bm{0}_{n_a \times n - n_a} \\  \bm{0}_{n - n_a \times n_a} & \bm{I}_{n - n_a}\end{bmatrix} \bm{q} \, ,
\end{equation}
where $\bm{g}: \mathbb{R}^{n} \rightarrow \mathbb{R}^{n_a}$ is the coordinate transformation in the actuated coordinates $\bm{\theta}_a$.
For continuum robots modeled with \ac{CRT}, $\bm{g}\left(\bm{q}\right)$ has a clear physical interpretation. 
The actuated coordinates correspond to the lengths of the individual actuators, such as
\begin{equation} \label{eq:theta_a}
    \bm{g}\left(\bm{q}\right) = \begin{bmatrix}
        L_{\textnormal{a}, 1} \\
        L_{\textnormal{a}, 2} \\
        \vdots \\
        L_{\textnormal{a}, n_a}
    \end{bmatrix} = \left(\int_{0}^{L}\bm{B}_{\bm{\tau}, i}^{\top} \left(\bm{\xi} + \begin{bmatrix}
        \bm{0} \\ \bm{d}'_i
    \end{bmatrix}\right) \, \textnormal{d}s \right)^{n_a}_{i = 1} \, ,
\end{equation}
where $L_{\textnormal{a}, i} \in \mathbb{R}$ denotes the length of the $i$-th actuator.

The change of coordinate \eqref{eq:theta_a} is significant because it associates a subset of the generalized coordinates with a physically meaningful quantity that can be easily measured \cite{pustina2024input}. 
For example, for a tendon-driven \ac{CSM}, the actuated variables $\bm{\theta}_{a}$ can be directly obtained using encoders.

In these collocated coordinates, the \acp{EoM} take the partially decoupled form, such as
\begin{equation}
    \bm{M}_{\bm{\theta}} \ddot{\bm{\theta}} + \bm{C}_{\bm{\theta}} \dot{\bm{\theta}} + \bm{G}_{\bm{\theta}} + \bm{K}_{\bm{\theta}} \bm{\theta} + \bm{D}_{\bm{\theta}} \dot{\bm{\theta}} = \begin{bmatrix}
        \bm{I}_{n_a} \\ \bm{0}
    \end{bmatrix} \bm{\tau} \, ,
\end{equation}
where $\bm{M}_{\bm{\theta}}, \, \bm{C}_{\bm{\theta}}, \, \bm{G}_{\bm{\theta}}, \, \bm{K}_{\bm{\theta}},$ and $\bm{D}_{\bm{\theta}}$ are the inertial, Coriolis, gravitational, stiffness, and damping matrices, respectively, expressed in the collocated coordinates $\bm{\theta}$.

Adopting this change of coordinates is promising for facilitating the transfer of control methods developed for underactuated rigid systems. For instance, \cite{pustina2024input} demonstrates the application of classical controllers—such as PD with gravity and elasticity compensation \cite{borja2022energy, borja2022role} and its variant P-satI-D \cite{pustina2022p}, a PID controller with integral saturation—directly to the collocated variables.
Finally, this formulation not only enhances the understanding of the system dynamics and simplifies control design, but also enables the design of linear controllers in the collocated coordinates without approximations. Notably, the control laws remain nonlinear due to the inherent nonlinearity of the mapping $\bm{g}\left(\bm{q}\right)$.

\subsection{Control by Model Inversion}
The most direct approach to controlling a continuum robot is through its kinematic model. This strategy, known as \ac{IK}, computes the actuation inputs required to drive the robot’s tip to a desired pose. Early studies demonstrated effective planar trajectory tracking and vibration suppression using simplified models such as \ac{EBRT} \cite{ataka2020model, gravagne2003Large}. To better handle real hardware complexities, these models were extended with online observers, which compensate for phenomena such as variable stiffness caused by changes in inflation pressure \cite{ataka2020model}.

To achieve higher accuracy and address complex spatial deformations, recent work shifted towards the more comprehensive \ac{CRT}, enabling advanced capabilities such as micrometer-level positioning in parallel continuum robots \cite{mauze2021micrometer} and multi-point orientation control along a manipulator’s body \cite{richter2021multi}. Since the accuracy of \ac{CRT} may increase the computational cost, limiting real-time use, researchers considered quasi-static formulations and actuator nonlinearity compensation to improve efficiency \cite{campisano2021closed}. 
These kinematic models are continuously being generalized to accommodate more complex physical designs, including manipulators with arbitrary, discontinuous, or overlapping actuator routings \cite{renda2022geometrically}.

While kinematic models determine the robot’s shape, dynamic models are required for high-speed maneuvers and interactions with the environment. Controllers based on \ac{ID} account for forces such as inertia, gravity, and external contacts. Such methods achieved full shape control, with some approaches combining geometric methods with \ac{EBRT} to balance accuracy and computational speed \cite{mbakop2021inverse}. 
Because dynamic models are computationally demanding, recent work explored decentralized control architectures that distribute computation across modules, improving real-time feasibility \cite{doroudchi2021configuration}.

\subsection{Managing Nonlinearities and Under-actuation}

A central challenge in continuum robotics arises from the strong nonlinearities and inherent under-actuation of these systems. \ac{FL} is a powerful technique that cancels nonlinear terms, transforming the system dynamics into a simpler, linear form. \ac{FL} was successfully applied in cascade control architectures to compensate for gravity and bending effects, thereby improving tracking performance \cite{falkenhahn2016dynamic}. However, its effectiveness depends heavily on model accuracy. To address this limitation, \cite{rucker2022task} proposed a hybrid approach that combines an inner-loop \ac{FL} controller with a robust outer-loop strategy such as \ac{SMC}, enabling robustness against model uncertainties \cite{rucker2022task}. Practical aspects also matter: research shows that the choice of discretization scheme can significantly influence numerical robustness, computational cost, and accuracy \cite{caradonna2024model}.

An alternative approach is the \ac{ESC}. Instead of canceling dynamics, \ac{ESC} reshapes the system’s potential energy such that the desired configuration becomes a stable energetic minimum. By reformulating the dynamics in a Port–Hamiltonian framework, this method can formally guarantee stability and passivity, allowing control of both under-actuated and hyper-redundant systems \cite{caasenbrood2021energy, caasenbrood2022energy}. \ac{ESC} often produces smooth, natural motions that exploit the robot’s intrinsic compliance. It was applied even to complex bio-inspired systems such as octopus tentacles, where optimal control was used to identify shaping laws for advanced models that include non-rigid cross-sections \cite{chang2023energy}. 

Discrete-time control is necessary for practical implementations, even though most of the \ac{ESC} literature concentrates on continuous-time models. Structure-preserving numerical techniques, such as Lie group variational integrators, are used to maintain the stability and passivity guarantees in a discretized system \cite{tiwari2023discrete}. These integrators allow for efficient stabilization of flexible beams while preserving the geometric structure of the manipulator's configuration space.

\subsection{Robust and Optimal Control}

Robust controllers for continuum robots were designed to mitigate model uncertainties. \ac{SMC} is a widely adopted strategy for ensuring robustness against model uncertainties and external disturbances. Adaptive variants of \ac{SMC} were developed to handle limitations such as actuator saturation while guaranteeing marginal stability \cite{alqumsan2019robust}. A classical drawback of \ac{SMC} is the high-frequency ``chattering” in the control signal. This issue was mitigated by integrating fuzzy logic to smooth the discontinuous control law, leading to smoother actuation profiles \cite{li2021modeling}. The methodology continues to evolve with advanced variants, such as terminal \ac{SMC}, which achieves faster finite-time convergence. It was combined with algorithms for high-level tasks such as obstacle avoidance \cite{mishra2023trajectory}.

When performance is critical and constraints must be explicitly respected, \ac{MPC} is the method of choice. \ac{MPC} solves a finite-horizon optimization problem at each time step, enabling it to naturally enforce input bounds while optimizing complex objectives. This capability makes it particularly suitable for medical applications, such as catheter control, where the robot must apply a constant force to tissue while tracking a trajectory \cite{soltani2017soft}. \ac{MPC} is also highly effective for motion planning, as it can generate the complex, oscillatory actuation signals required for behaviors like fish-like swimming, even under disturbances \cite{barbosa2023motion}. However, the limitation of \ac{MPC} remains its computational cost, which often necessitates model simplifications (e.g., linearization) for real-time implementation.


Beyond traditional methods, emerging paradigms include optimal control for bio-inspired motions \cite{wang2021optimal}, decentralized robust control for modular architectures \cite{doroudchi2018decentralized}, and synergetic control \cite{hashemi2023robust}, which has recently shown robustness to disturbances in continuum robots control.


\begin{table*}[t]
\caption{Model-based controllers using rod models.}
\label{tab:mb-controllers-table}
\small\sf\centering
\resizebox{0.9\textwidth}{!}{  
\begin{tabular}{ccccc}
\toprule
\textbf{Controller} & \textbf{References} & \textbf{Rod Model} & \textbf{Task} & \textbf{Validation} \\
\midrule
Inverse Kinematics & \begin{tabular}[c]{ccccc} \cite{ataka2020model, ataka2020observer} \\ \cite{mauze2021micrometer} \\ \cite{richter2021multi} \\ \cite{campisano2021closed} \\ \cite{renda2022geometrically} \end{tabular} & \begin{tabular}[c]{ccccc} \ac{EBRT} \\ \ac{CRT} \\ \ac{CRT} \\ \ac{CRT} \\ \ac{CRT} (GVS) \end{tabular} & \begin{tabular}[c]{ccccc} Tip Pose \ac{TT} \\ Tip Position \ac{TT} \\ \ac{MP} Orientation Reg. \\ Tip Pose \ac{TT} \\ \ac{MP} Pose \ac{TT} \end{tabular}  & \begin{tabular}[c]{ccccc} Experimental \\ Experimental \\ Experimental \\ Experimental \\ Numerical \end{tabular} \\
\midrule
Inverse Dynamics & \begin{tabular}[c]{cc} \cite{mbakop2021inverse} \\ \cite{doroudchi2021configuration} \end{tabular} & \begin{tabular}[c]{cc} \ac{EBRT} + PH \\ \ac{CRT} \end{tabular} & \begin{tabular}[c]{cc} Shape Regulation \\ Shape \ac{TT} \end{tabular} & \begin{tabular}[c]{cc} Experimental \\ Numerical \end{tabular} \\
\midrule
Feedback Linearization & \begin{tabular}[c]{cc} \cite{rucker2022task} \\ \cite{caradonna2024model} \end{tabular} & \begin{tabular}[c]{cc} KRT \\ \ac{CRT} (GVS) \end{tabular} & \begin{tabular}[c]{cc} Tip Position \ac{TT} \\ Pick and Place \end{tabular} & \begin{tabular}[c]{cc} Numerical \\ Numerical \end{tabular} \\
\midrule
Energy-Shaping & \begin{tabular}[c]{ccc} \cite{caasenbrood2021energy, caasenbrood2022energy} \\ \cite{chang2023energy} \\ \cite{tiwari2023discrete} \end{tabular} & \begin{tabular}[c]{ccc} \ac{CRT} (GVS) \\ \ac{CRT} (DER) \\ Planar \ac{CRT} \end{tabular} & \begin{tabular}[c]{ccc} Tip Position Reg. \\ Reaching, Grasping \\ Shape Reg. \end{tabular} & \begin{tabular}[c]{ccc} Numerical \\ Numerical \\ Numerical \end{tabular} \\
\midrule
Sliding Mode & \begin{tabular}[c]{ccc} \cite{alqumsan2019robust} \\ \cite{li2021modeling} \\ \cite{mishra2023trajectory} \end{tabular} & \begin{tabular}[c]{ccc} \ac{CRT} \\ \ac{CRT} \\ \ac{CRT} + Hyst. \end{tabular} & \begin{tabular}[c]{ccc} Tip Pose \ac{TT} \\ Tip Position \ac{TT} \\ Tip Position \ac{TT} \end{tabular} & \begin{tabular}[c]{ccc} Numerical \\ Experimental \\ Numerical \end{tabular} \\
\midrule
Model Predictive Control & \begin{tabular}[c]{cc} \cite{soltani2017soft} \\ \cite{barbosa2023motion} \end{tabular} & \begin{tabular}[c]{cc} Static \ac{CRT} \\ \ac{EBRT} \end{tabular} & \begin{tabular}[c]{cc} Tip Position \ac{TT} \\ Tip Position \ac{TT} \end{tabular} & \begin{tabular}[c]{cc} Experimental \\ Numerical \end{tabular} \\
\midrule
Open-Loop Optimal Controller & \cite{wang2021optimal} & \ac{CRT} (DER) & Reaching, Fetching & Numerical \\
\midrule
$H_\infty$ & \cite{doroudchi2018decentralized} & \ac{EBRT} & Shape Regulation & Numerical \\
\midrule
Synergetic Control & \cite{hashemi2023robust} & \ac{CRT} & Tip Position \ac{TT} & Numerical \\
\bottomrule
\end{tabular}
}
\end{table*}

\subsection{Shape/State Estimation} 
\label{mb_controllers:shape_estimation}
    A general \ac{MB} shape controller requires feedback in terms of configuration vector $\bm{q}$ and its time derivative $\dot{\bm{q}}$ (\autoref{fig:mb_scheme}). Depending on the spatial discretization, the vector $\bm{q}$ could be difficult or impossible to measure. To tackle this issue, a general \ac{MB} control framework requires an Estimator, which maps the measures $\bm{y}$ to the estimated state $\hat{\bm{q}} \, , \, \dot{\hat{\bm{q}}}$. For continuum and soft robots, this problem is referred to as Shape/State Estimation or Shape/State Reconstruction. Shape Estimation methods can use geometrical models or Computer Vision algorithms \cite{bezawada2022shape, albeladi2021vision}. However, the following works show that incorporating prior knowledge of rod-based models can significantly enhance performance.
The pioneering work of \cite{trivedi2014model} suggests three approaches for Shape Estimation, based on \ac{CRT}: (i) mounting load cells at the base, (ii) employing cable encoders, and (iii) mounting an inclinometer at the end of each piece. A maximum tip position error of $3\%L$ is observed in simulations. All three approaches were then experimentally validated on the OctArm VI \cite{grissom2006design}.
Another study by \cite{donat2021real} proposes a real-time shape-estimation method based on the force-torque measured at the basis of the tubes for a \ac{CTR}. It extends a shape estimation algorithm for elastic Kirchhoff rods. They modeled a \ac{CTR} combining planar \ac{PCC} segments lying on different equilibrium planes. The approach is evaluated with single and two combined additively manufactured tubes at high frequency, achieving a mean deviation of $2 - 5$ mm along the tube.
Differently, \cite{lilge2022continuum} developed a model for shape and strain estimation of a continuum cable-driven robot using a Gaussian Process regression to estimate continuous trajectories in $SE(3)$. The idea involves substituting (i) time $t$ with arc-length $s$ and (ii) kinematic and dynamic laws based on \ac{CRT}. This method efficiently estimates the robot's shape using noisy measurements from sensors such as strain gauges, tracking coils, and an external camera. The real-world experiments provide excellent performances, with an average tip pose error of $3.3$ mm and $0.035^\circ$.
In parallel, \cite{aloi2022estimating} developed a model-based framework for estimating distributed contact forces directly from shape measurements. The method involved fitting a Cosserat rod kinetostatic model with Gaussian load parameterization and solving it via an Extended Kalman Filter, enabling real-time, sensorless force estimation with sub-Newton accuracy. State estimation under uncertainty was also addressed in reconfigurable parallel continuum systems \cite{anderson2017continuum}, applying Cosserat rod modeling, multi-point constraints, and advanced filtering to improve tip accuracy and enable load inference. \cite{ferguson2024unified} presents a surgical continuum manipulator with integrated FBG sensing for real-time 3D shape reconstruction, enabling sub-millimeter navigation without radiation in complex surgical settings. \cite{shi2016shape} provided a comprehensive review of shape-sensing modalities for surgical continuum robots, highlighting complementary strengths of electromagnetic, optical, FBG, and vision-based methods.
On the control side, underactuated discrete rod models were leveraged to achieve precise task-space trajectory tracking through \ac{FL} and robust sliding-mode strategies, all while relying on minimal sensing \cite{rucker2022task}.
In addition, \cite{yousefi2023model} propose an optimization-based method that simultaneously estimates the shape and the forces acting on a continuum robot by employing the quasi-static \ac{CRT}. Magnetic localization determines the position of multiple robot points. The method estimates the robot's shape and force in a wide range of conditions validated experimentally, with and without knowledge of the contact positions.
Recently, \cite{zheng2024full,zheng2024estimating} proposed the Cosserat theoretic boundary observer, a mechanics-based dynamic state estimation algorithm to recover the infinite-dimensional robot states by measuring the tip velocity twist and the tip pose.
Finally, \cite{feliu2024dynamic} presents a novel method for estimating the dynamic state (pose and velocity) of tendon-actuated \acp{CSM}. The approach uniquely combines a \ac{GVS} Cosserat rod model with a nonlinear State-Dependent Riccati Equation observer. A key advantage of this technique is its ability to estimate the state using only internal sensor data, such as tendon displacements and actuator forces. The method was validated both in simulation and on a prototype, achieving an average dynamic tip position error of 1.79 cm.

\section{Learning-based Controllers for Rod Models}
\label{sec:learning_based_controllers}
\ac{ML} methods for controlling \acp{CSM} are becoming a trend \cite{george2018control, wang2021survey, wang2022control, falotico2024learning}. Learning-based controllers can leverage the physics of rod models to derive control policies $\pi(\cdot \ ; \ \bm{w})$ from experience. Such policies establish end-to-end mappings between desired poses $\bm{g}_d$, sensor measurements $\bm{y}$, and motor commands $\bm{\tau}$:
\begin{equation}
\bm{\tau} = \pi(\bm{g}_d, \bm{y}; \ \bm{w}),
\label{eq:learning-based_controller}
\end{equation}
where $\bm{w}$ represents the weights of an \acp{ANN}. By integrating physics-based data, these methods can improve adaptability and robustness in automation tasks that involve compliant interactions with the environment. Below, we review learning-based policies for \ac{SL} and \ac{RL}. We adopt a chronological perspective to highlight the evolution of this growing field and its impact on robotics and automation.

\subsection{Supervised Learning}
\ac{SL} in simulation is a viable approach to address simple tasks with soft robots. It relies on pseudo-random motion data of actuations $\bm{\tau}$ and corresponding task space $\bm{g}$ to train an inverse model \eqref{eq:learning-based_controller}. 
For instance, \cite{thuruthel2017learning} developed dynamic models of \acp{CSM} using \acp{RNN} and presented a trajectory optimization method for task space control. They validated the controller on a Cosserat rod with \ac{PCS} parameterization and a pneumatic \ac{CSM}.
Similarly, \cite{wang2022data} developed a closed-loop controller for continuum robots using \acp{RNN} to approximate forward and inverse dynamics. They extended the model in \cite{till2019real} to account for spine compression and generate motion data for training. A non-parametric Gaussian process regression compensates for discrepancies between the real robot and the \ac{RNN}, while a hybrid controller alternates between two stand-alone policies. The approach was validated on reaching and tracking tasks with a tendon-driven continuum robot, showing improved accuracy by combining simulated and experimental data.
Recently, generative adversarial networks achieved domain translation of a pose controller across environments with different medium viscosities, supported by a Cosserat rod model \cite{kushawaha2025domain}. While combining continuum and generative models for adaptive soft robotic automation is promising, the domain-translation model required substantial computing, and the policy only solved tracking tasks in simulation.

\begin{figure}[t]
    \centering
    \includegraphics[width=0.9\columnwidth]{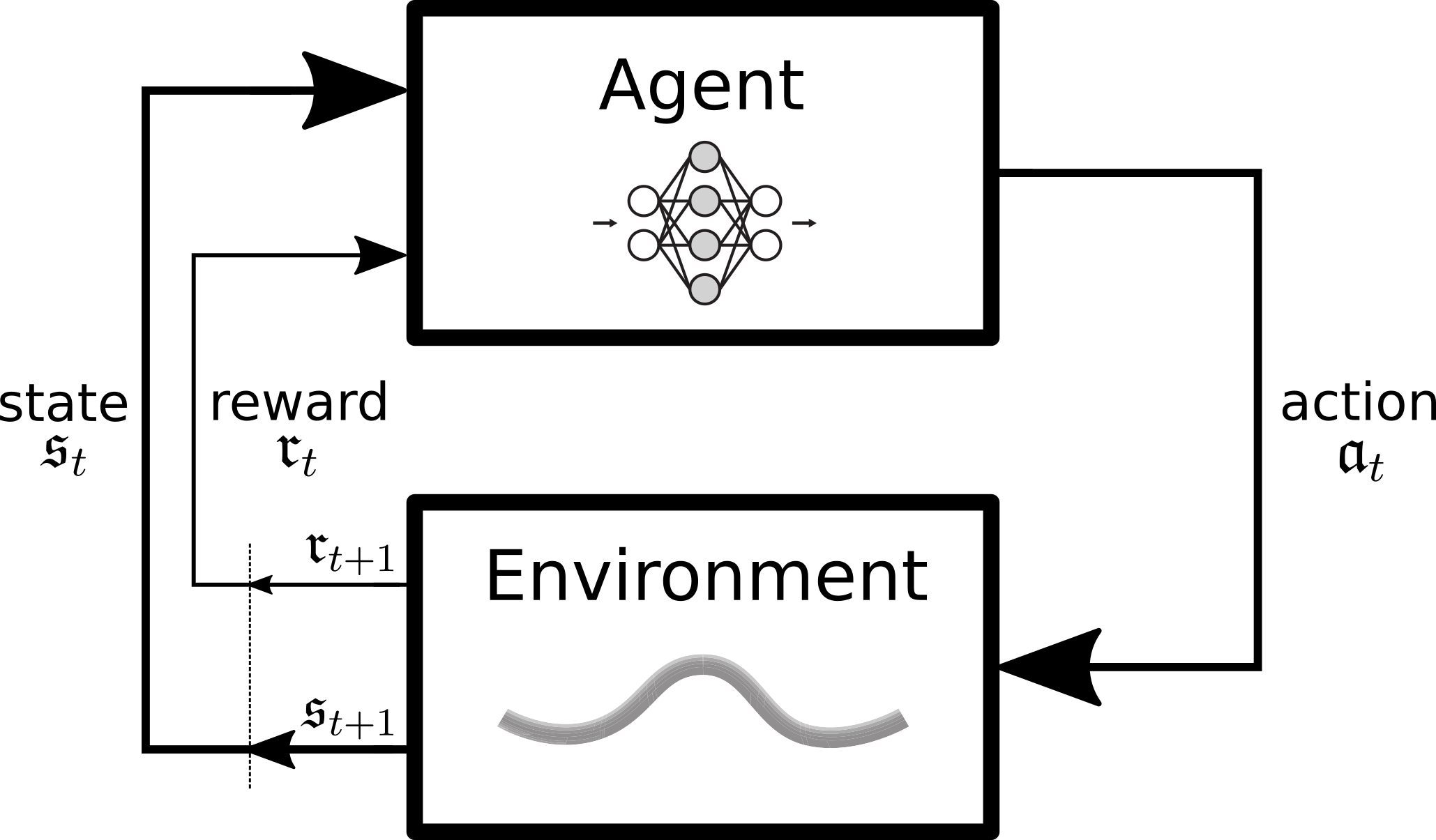}
    \caption{Soft robot control combing \ac{RL} and rod models. The agent employs an \ac{ANN} as controller, while rod models govern the environment dynamics.}
    \label{fig:rl-schema-rod}
\end{figure}

\subsection{Reinforcement Learning}
\textbf{Overview}: \ac{RL} is a versatile framework for sequential decision-making, adaptive control, and robotic automation \cite{sutton2018reinforcement}. The two fundamental objects of an \ac{RL} problem are the \textit{agent} and the \textit{environment}. In soft robotic control, the agent is a learning-based controller based on \acp{ANN}, whereas the environment is everything outside the agent, comprising the soft robot and other objects. \autoref{fig:rl-schema-rod} depicts the agent-environment interaction highlighting the role of \acp{ANN} and rod models.
The agent and the environment interact at discrete time steps $t$. At every step, the agent observes the (possibly partial) state $s_t$ of the environment, including sensor readings and task information. The agent acts according to a policy $\pi$ to select actions $a_t$ taking the form of low-level controls $\bm{\tau}$ like torques or pressures. A policy can be a \textit{deterministic} mapping from states to actions, $a_t = \pi(s_t)$. Alternatively, a policy can be \textit{stochastic}, mapping a state to a distribution over actions $a_t \sim \pi(\cdot | s_t)$.
As a result of taking an action, the environment transitions to a new state $s_{t+1}$ according to the rod dynamics. Then, the agent perceives a scalar reward signal $r_{t+1}$ from the environment, indicating the goodness of the current state specifying the control task. The goal of the agent is to maximize its \textit{return}, the cumulative sum of rewards:
\begin{equation}
G_t = \sum_{k=0}^\infty \gamma^k \mathfrak{r}_{t+1+k}, 
\end{equation}
where $\gamma \in (0,1)$ is a discount factor. To sum up, \ac{RL} aims to find an \textit{optimal} policy $\pi^*$ that maximizes the expected return
\begin{equation}
\pi^* = \arg \max_{\pi} \mathbb{E}_{\pi} \left[ G_0 \right].
\end{equation}

Below and in \autoref{tab:rl-controllers-table}, we review the state-of-the-art \ac{RL} policies for \acp{CSM} trained on rod models.

\textbf{Survey of RL policies}:
The earliest contribution was presented by \textit{Satheeshbabu et al.} \cite{satheeshbabu2019open}, where they applied \ac{DQL} with experience replay. \ac{DQL} can cope with the challenging state-action space of soft robots by employing an \ac{ANN} $\hat{Q}(\mathfrak{s}, \mathfrak{a}; \bm{w})$ to approximate the \textit{value} function of state-action pairs \cite{mnih2013playing}, which is defined as

\begin{equation}
    Q_\pi(\mathfrak{s}, \mathfrak{a}) \coloneqq \mathbb{E}_\pi \left[ G_t \, | \, \mathfrak{s}, \mathfrak{a} \right].
\end{equation}
Then, a deterministic policy can leverage $\hat{Q}$ to select the best action in a state as $a = \pi(s) = \arg \max_a \hat{Q} (s, a; \bm{w})$.
The work presented an open-loop policy for the quasi-static position control of a pneumatic \ac{CSM} that can bend and twist. The policy $\pi$ was trained on a static Cosserat rod simulation and validated both in simulation and on the physical platform, subject to various external loads. 
The work was extended by increasing the dexterity of the \ac{CSM} and developing a closed-loop policy, still for precise quasi-static positioning, via \ac{DDPG} \cite{satheeshbabu2020continuous}. They investigated the robustness of the control policy to loading and workspace discontinuity, and deployed the policy on hardware.
The \ac{DDPG} algorithm also derived a closed-loop controller for the quasi-static positioning of a pneumatic \ac{CSM} modeled with the \ac{KRT} \cite{uppalapati2020berry}. Interestingly, the \ac{CSM} was integrated into a mobile platform including a rigid arm and a sensorized gripper. The system was teleoperated and validated in the agricultural task of picking berries using different maneuvering strategies.

The potential of continuum mechanical models is fully exploited in automatic physical interaction tasks.
\textit{Naughton et al.} \cite{naughton2021elastica} effectively applied several deep \ac{RL} algorithms to train various control policies. They explored point reaching, trajectory tracking, and maneuvering through structured and unstructured obstacles. The controllers were learned and tested in simulation using a synthetic \ac{CSM} based on a dynamic \ac{CRT}. The rod was actuated by applying distributed internal torques, modeled via splines characterized by control points and vanishing values at the rod ends. The work showcased a captivating interaction between the rod and the environment. For example, the maneuvering task required the rod free end $\bm{r}(L)$ to reach a position $\bm{x}_{d}$ behind fixed obstacles. The task was solved with a reward function that only considered the distance to the target and a penalty term for numerical errors
\begin{equation}
    \text{reward} = 
    \begin{cases}
        \text{penalty}, & \text{numerical errors}\\
        ||\bm{x}_{d} - \bm{r}(L) ||, & \text{otherwise}.
    \end{cases}
\end{equation}
Interestingly, the rod learned to navigate and interact with the obstacles without an explicit reward term.

The generalization capabilities of learned controllers are vital for successful real-world deployment. In this regard, \textit{Alessi et al.} \cite{alessi2023learning} applied \ac{PPO} to train a closed-loop position control policy for dynamic trajectory tracking with a pneumatic \ac{CSM}. The controller was trained leveraging a dynamic Cosserat rod model of the soft robot \cite{alessi2023ablation}. Then, four simulation tests evaluated how the policy generalized to new observations, dynamics, and tasks. The experiments included tracking new trajectories subject to unknown external forces $\bm{\mathcal{F}}_e(L)$ or using different material properties (e.g., $E$). The policy generalized well within some boundaries and also transferred without retraining to intercept a moving object. 
Subsequently, \cite{alessi2024pushing} extended the simulation environment to account for the dynamics of a dexterous \ac{CSM} and its interaction with objects. The proposed closed-loop pose/force controller enabled dynamic pushing with \acp{CSM} in the real world. To mitigate the significant sim-to-real gap in soft robotics, the authors introduced a novel adaptation of \ac{DR} \cite{zhao2020sim} emphasizing the soft material properties. This work demonstrated the first sim-to-real transfer of a rod-based manipulation policy, paving the way for soft robotic automation and physical interaction tasks.

\begin{table*}[t]
\caption{Deep reinforcement learning controllers using rod models.}
\label{tab:rl-controllers-table}
\small\sf\centering
\resizebox{0.9\textwidth}{!}{  
\begin{tabular}{@{}ccccc@{}}
\toprule
\textbf{Controller} &
\textbf{References} &
  \begin{tabular}[c]{@{}c@{}}\textbf{Rod Model}\\ (dynamics)\end{tabular} &
  \textbf{Task} &
  \begin{tabular}[c]{@{}c@{}}\textbf{Sim-to-real}\\ (technique)\end{tabular}\\ 
  \midrule
Deep Q-Learning &
\cite{satheeshbabu2019open} & 
CRT (\xmark) & 
Tip Position \ac{TT} (way-point) & 
\cmark (None) \\
\midrule
\multirow{2}{*}{Deep Deterministic Policy Gradient}  & 
\cite{satheeshbabu2020continuous} & 
CRT (\xmark) &
Tip Position \ac{TT} (way-point) & 
\cmark (None)\\
  &
\cite{uppalapati2020berry} &
KRT (\xmark) & 
Tip Position \ac{TT} (way-point) &
\cmark (None) \\
\midrule
Trust Region Policy Optimization &
  \cite{naughton2021elastica} &
  CRT (\cmark) &
  Tip Pose \ac{TT}, Obstacle maneuvering &
  \xmark \\
  \midrule
\multirow{2}{*}{Proximal Policy Optimization} &
\cite{alessi2023learning} &
  \ac{CRT} (\cmark) &
  Tip Position \ac{TT}, Object interception &
  \xmark \\ 
  &
  \cite{alessi2024pushing} &
  \ac{CRT} (\cmark) &
  Pushing (Tip Pose/Force)
  &
  \cmark (\ac{DR})\\
  \bottomrule
\end{tabular}
}
\end{table*}

\section{Discussion and Emerging Challenges}
\label{sec:discussion_and_emerging_challenges}
\subsection{On the Rod Theories}
Following our mathematical treatment and review of rod models, we briefly reevaluate the strengths and limitations of the various rod theories within soft robotics. The Euler-Bernoulli model is simple, computationally light, and effective for slender, stiff structures subject to small bending deflections. Its nonlinear version makes it suitable for large deformations. However, it remains a planar model and neglects shear as the cross-sections remain perpendicular to the centerline. This assumption makes it unsuitable for modeling interactions with the environment in soft robotics. Nonetheless, some works extended its applicability beyond bending by coupling it with other models. The Timoshenko model improves on this by including shear deformation. However, Timoshenko is less suited to nonlinear actuator paths that induce torsion. To this end, the Kirchhoff–Love theory handles large rotations and displacements but assumes the rod is inextensible and unshearable. This assumption limits its ability to model extensible pneumatic actuators. Finally, the Cosserat-Reissner model and its extensions are the most general and suited for continuum soft robots. They can handle large deformations, shear, torsion, and stretching. However, they require more computational effort. Overall, each rod theory has its strengths and weaknesses. The choice depends on the system’s behavior and the level of detail needed. We suggest opting for a Cosserat rod model and tuning the space and time discretization to match the desired tradeoff between efficiency and accuracy.

\subsection{On the Deformations}
Researchers developed numerous rod-based models by adapting classical rod theories to soft robotics. The reviewed models capture different combinations of strain modes, giving rise to nine deformation classes.

The most common class in literature is \textit{Bend}, with about a third of the examined papers. This popularity is due to the propensity of slender bodies to bend. In fact, due to the assumption of $L \gg D$, with $D$ cross-section diameter, the bending stiffness turns out smaller than the stretching or shear ones, easing the bending deformation. In addition, most used actuator paths are linear, which excite predominantly bending modes. These reasons justify the approximation of considering only the bending mode. Furthermore, we can observe that most models in the Bend class leverage the \ac{EBRT}. However, as summarized in \autoref{tab:unified_model_table}, several models employed the general \ac{CRT}. While this may constitute a computational surplus for simple bending structures, the expressive power of Cosserat models becomes valuable when the model is used in physical interaction scenarios, for instance, for policy learning or model-based control design.

The next class is \textit{All Deformation}, with about a quarter of the examined papers. As expected, only models based on the \ac{CRT} achieve all deformations (\autoref{tab:unified_model_table}). This group of models received a deep study in the last decade, thanks to the rising interest in physical interaction with unstructured environments where all strain modes are significant. Indeed, these models are useful to describe accurately the deformations of slender robots caused by friction (e.g., anisotropic friction for locomotors) or by different mediums (e.g., water). In addition, the most common actuation sources (i.e., pneumatic and cables) excite elongation and compression, causing a significant change in the dimension workspace. 

The third class accounting for about a fifth of the investigated studies is \textit{Bend \& Twist}. These models mainly describe robots in which the unshearability and the unstretchability approximations hold (\autoref{tab:unified_model_table}). Including the twisting deformation allows the representation of backbones with generic \ac{3D} curves. However, the twisting deformation can be excited only by specific actuator paths (e.g., helicoidal cables) or asymmetric actuators.

Another significant class is \textit{Bend \& Stretch} thanks to the wide diffusion of linear extensible soft pneumatic actuators and tendon-driven robots. Such actuators extend or compress upon solicitation, which induces the continuum body to stretch and bend thanks to their radial disposition on the cross-section. 

The remaining five categories are less explored. While the \textit{Bend, Twist \& Stretch} is a particular case of the \textit{All Deformations} class, the other classes include fewer works. The lack of work is due to the difficulty of exciting these specific strain modes with the current actuation technologies or the small number of applications in which the assumptions are valid. The fact that fewer studies explored shearing per se could be due to the tendency to consider shearing an undesirable or negligible deformation. However, shearing could be significant when there is a prominent environment interaction, as observed in peristaltic locomotion \cite{hemingway2021continuous} or complex manipulation \cite{grazioso2019geometrically}. Shearing also becomes important when the aspect ratio of the \ac{CSM} becomes less slender and thicker.

\subsection{On the Spatial Discretization}
A characterizing feature of rod models is the \textit{spatial discretization} method, which significantly impacts the model's accuracy and complexity. For simulation and control purposes, the discrete model must be computationally efficient and as accurate as possible. In the analyzed literature, two primary discretization approaches are commonly employed: \ac{DER} and Strain Parameterization.

The \ac{DER} method discretizes the rod’s length into nodes connected by rigid links, representing the rod \ac{CSM} as a set of discrete poses. The accuracy of the solution relative to the continuous rod depends on the number of nodes. In contrast, Strain Parameterization discretizes the rod in the configuration space, constraining the robot’s backbone to a subset of admissible spatial curves. The latter approach permits the neglect of specific strain modes without imposing additional constraints on the \acp{EoM}, as required in the \ac{DER} formulation, thereby enhancing computational efficiency. Furthermore, Strain Parameterization naturally allows the forward dynamics to be expressed in Lagrangian form, which is particularly advantageous for control design. Both approaches were widely used and integrated into dedicated simulators, such as SoRoSim \cite{mathew2022sorosim} for Strain Parameterization and PyElastica \cite{naughton2021elastica} for \ac{DER}.

Straddling geometric and mechanical models, some examined works proposed to fit discrete points with functions, such as \ac{PH} or Euler's spirals, which is useful, especially for Shape Estimation.

\subsection{On the Identification and Validation}
Identification and validation of rod models play a crucial role in accurately reproducing the experimental robot behaviors.
Despite the high computational cost, it emerged that detailed \ac{FEM} simulations served a dual purpose: (i) to assess the correctness of the rod model, and (ii) to estimate the unknown parameters. 
In the absence of precise \ac{FEM} models, lumped-mass models with a high number of \ac{DoFs} were employed. Otherwise, researchers resort to model optimization to match experimental data.
In particular, when the robot morphology is not uniform, researchers usually estimate geometrical properties (e.g., cross-sectional area $A(s)$) and material characteristics (e.g., elastic modulus $E$, damping coefficient $\beta$). The experiments conducted typically feature the robot at different base orientations subject to external loads $\bm{\mathcal{F}}_e(s)$ or at the tip. Most rod models are identified and validated through motion data collected with predefined actuation profiles $\bm{\tau}$. However, rigorous mechanical tests are still needed to yield more accurate models, especially for unstructured physical interactions with the environment. Indeed, despite a careful model calibration on motion data, \cite{alessi2023ablation, alessi2024pushing} observed real-world buckling during a manipulation task that was not present in simulation.
Concerning performance metrics, rod models predicted experimental tip positions with Euclidean errors in the range $1-10\%L$. However, only a few studies assessed the angular errors, which play a role in dexterous manipulators. In summary, the increasing use of experimental data reflects the demand to bridge the sim-to-real gap for developing more accurate controllers. 

\subsection{On the Impact of Materials and Manufacturing}
Most of the analyzed rod models employed continuum and soft robots built with diverse materials, including traditional elastomers (e.g., silicone, rubber) or other polymers (e.g., \acp{SMA}, hydrogels). Each material offers distinct properties, influencing the robot's compliance, flexibility, and functionality. However, effectively modeling the nonlinear behavior inherent in soft materials using rod theories presents challenges. For instance, some combinations of materials and actuation mechanisms cause notable changes in cross-section areas or volume, which may render the rod-like assumption overly restrictive. Nonetheless, an extension of the \ac{CRT} considering geometrical rescaling \eqref{gazzola_rescaling} could approximate complex dynamics \eqref{gazzola_dynamics}.

The viscoelastic constitutive law of the material impacts significantly on the accuracy. In the classic rod theories, this relation is supposed linear, as already shown in \eqref{viscoelastic_law_cosserat}. However, especially in the case of pneumatic actuation, the linear constitutive law is no longer valid. To tackle this issue, many analyzed works presented models with a nonlinear constitutive law, using hyperelastic model such as Odgen.

Manufacturing techniques provided continuum and soft robots with unprecedented dexterity \cite{wallin20183d}. However, manufacturing uncertainties may impact model accuracy. Recent rod models addressed manufacturing uncertainties considering variations in material \cite{alessi2023ablation} and geometric properties \cite{eugster2022soft}. 

Research gaps persist despite these advancements. First, capturing hysteresis---a prevalent phenomenon in soft materials that affects the accuracy and predictability of the robot's behavior---remains largely underexplored in rod models. Second, incorporating self-healing materials can potentially enhance the durability and longevity of soft robots \cite{terryn2017self}. However, integrating the dynamic restoring process into rod models should still be investigated. Filling these gaps could advance the capabilities of continuum and soft robots, enabling their integration into diverse applications and real-world tasks. 

\subsection{On the Model-based Controllers}
The emergence of computationally efficient and accurate models enables researchers to develop model-based controllers that can accurately predict and react to nonlinear elastic rods.
We observed that controllers based on \ac{IK} are the most popular choice, as they achieve tip trajectory tracking tasks in real-world prototypes (Table~\ref{tab:mb-controllers-table}). Despite the simplified model, the feedback action can reject unmodeled effects, showing the effectiveness of that strategy. However, we should note that most IK-based controllers treat the system as fully actuated. This limits control to a smaller set of \ac{DoFs} than the robot is inherently capable of exhibiting.

A dynamic model is essential to exploit the properties of continuum robots. Most of the proposed controllers adopt well-known approaches from the under-actuation literature, such as \ac{FL}, \ac{ESC}, \ac{SMC}, or \ac{MPC}. However, only a few proposed controllers were validated experimentally. The reasons for neglecting the \textit{sim-to-real} comparison could be the lack of robustness or physical limits of the prototypes. While the Inverse Dynamics, \ac{FL}, and \ac{ESC} are sensitive to unmodelled effects and parametric uncertainties, the discontinuous output of the \ac{SMC} is difficult to implement with the most popular actuation technologies (e.g., tendons, pneumatic). Specifically, the actuators' internal dynamics may limit the effective system bandwidth, hence filtering the discontinuous output of the controller. The \ac{MPC} is a promising but not widely explored controller in the continuum robots literature, which could be attributed to the relatively high computational cost.

Two key aspects facilitate the design of controllers for \acp{CSM}: (i) the development of advanced discretization techniques and (ii) the tunability of model accuracy. The former enables the design of controllers with a finite number of state variables, allowing for the efficient computation of the robot’s kinematics and dynamics. Choosing appropriate discretization techniques is particularly critical for determining the minimum control frequency required to ensure numerical stability. Furthermore, the \ac{DER} technique captures cross-sectional deformations, which are especially relevant for modeling and controlling muscle-like structures in bio-inspired robotic systems \cite{chang2023energy}. 
This feature was recently integrated within the \ac{GVS} framework by \cite{sun2025real}, who proposed an extended \ac{CRT} formulation that models radial (in-plane) deformations through an additional strain variable. Their model also incorporates nonlinear hyperelastic and viscous effects, extending the \ac{GVS} approach to more complex, bio-inspired dynamic scenarios. Equally important is the ability to tune model accuracy by neglecting specific modes or reducing the number of \ac{DoFs}, as implemented in the Strain Parameterization. This flexibility allows for trade-offs between accuracy and computational complexity, enabling the controller to treat unmodeled dynamics as external disturbances. Such a strategy increases the control loop frequency by reducing computational load and enhances its overall robustness.

Regarding research gaps, we observed the lack of experimental validation for \ac{MB} controllers. Furthermore, there are few studies involving external forces due to interactions with the environment, particularly because of the complexity involved in modeling and estimating contact and friction forces. Due to the current lack of maturity of the controllers, the literature on rod-based planning \cite{mishra2023trajectory, bentley2023safer} is still sparse, although planning algorithms using geometrical approaches were recently proposed \cite{rao2024towards}. Future developments on model-based controllers could benefit from using \textit{physics-informed} \acp{ANN} to lessen the computational burden of complex expressions (e.g., the Coriolis matrix) or use them to compute the \acp{EoM} of the elastic rod, as already explored in \cite{lutter2023combining, liu2024physics}.
Another promising future direction is the application of the Collocated Form proposed in \cite{pustina2024input} to the design of more advanced controllers, particularly those inspired by the rigid underactuated robotics literature. This specific change of coordinates can simplify feedback implementation, since the actuator lengths are directly measurable and easier to estimate, as also exploited by \cite{feliu2024dynamic}. Furthermore, adopting this coordinate transformation can enhance \ac{IK}-based controllers by naturally accounting for the system’s under-actuation. Finally, differentiable simulators and libraries for rod models could be utilized for optimal control. Recently, analytical derivatives for the \ac{GVS} model were calculated in \cite{mathew2024analytical}, facilitating the implementation of model-based optimal controllers.

\balance
\subsection{On the Learning-based Controllers}
Learning-based controllers utilized the physics of rod models with a clear distinction between \ac{SL} and \ac{RL}. 

In \ac{SL}-based methods, rod models were only used as proxies to collect motion data under predefined open-loop policies. The motion data were employed to train \acp{ANN}, which efficiently emulated forward or inverse robot models, and replaced rods to facilitate the preliminary validation of control hypotheses. Although \ac{SL} was widely used in early soft robotic controllers, it limits the robot tasks to simple reaching and tracking due to the challenges in collecting supervised data for physical interaction tasks. For this reason, we believe \ac{SL} for rod models could fade away in the future.

Conversely, the role of rod models is central for developing complex deep \ac{RL} control policies to achieve unstructured physical interaction with \acp{CSM}. Deep \ac{RL} agents exploited computationally feasible rod models embedded in simulated environments to learn optimal actions $\bm{\tau}$ iteratively. Herein, \ac{RL} reduces the burden of collecting labeled datasets offline and unlocks new possibilities that are not fully explored yet. The control tasks addressed evolved from position control for quasi-static tracking to simulated pose control for obstacle maneuvering and pose/force control for dynamic pushing (Table~\ref{tab:rl-controllers-table}). For instance, initial policies limited the physical interaction to static payloads without addressing the sim-to-real gap \cite{satheeshbabu2020continuous}. Conversely, efforts in developing simulated environments \cite{naughton2021elastica} combined with careful reward engineering and sim-to-real techniques \cite{alessi2024pushing} facilitated recent advancements in physical interaction with \acp{CSM}. In summary, we believe the combination of rod models, deep \ac{RL} methods, and sim-to-real transfer techniques could advance the manipulation capabilities of \acp{CSM}.

All these works, however, relied on motion capture systems to track the robot positions. Indeed, reliable soft robotic proprioception is an open challenge \cite{pagliarani2025softtex}. Future developments could explore model-based \ac{RL}, where \textit{model-based} here refers to not only learning a policy but also learning a predictive model of the environment used for planning \cite{moerland2023model}. Alternative research lines can be borrowed from rigid robotics and tailored to the nonlinearities of continuum and soft robots \cite{ibarz2021train}.

\section{Conclusion}
\label{sec:conclusion}
This review paper explored the modeling and control of continuum and soft robots with rod theories. Our vertical literature survey spanned mathematical formulations of rod theories, rod-based models of continuum and soft robots, and control strategies with rod models. The review of the mathematical background facilitated comparison between different rod theory variants while highlighting connections to other models. We uncovered the versatility of rod theories through a comprehensive review of rod-based models that supported various studies, from bioengineering design to experimental validation of continuum and soft robots. We grouped the rod-based models in deformation classes to offer new perspectives on the evolution of the state of the art, particularly regarding the relationship between robot deformations and modeling choices. While advanced model-based control approaches were effective for tracking tasks, \ac{RL} coupled with sim-to-real strategies emerged as a promising approach for manipulation.

Despite remarkable advancements in integrating rod theories within soft robot models and controllers, limitations persist. For instance, modeling hysteresis effects of soft materials and comprehensive experimental validations in contact-rich scenarios remain challenging. Related to the control, addressing efficient sim-to-real policy transfer remains essential for robot deployment. While this paper focused on rod models for continuum and soft robot control, we believe the presented insights can broadly support future research in soft robotic automation, manipulation, and interaction with complex environments, and serve as a reference for developing continuum and soft robotic systems.


\appendices
\setcounter{equation}{0}
\numberwithin{equation}{section}
\section{Query for models and controllers}
\label{sec:Appendix-A}
The two Scopus advanced search queries used for the models and controllers are respectively:

\begin{itemize}
    \setlength\itemsep{1em}
    \item TITLE-ABS-KEY (``soft" \textbf{OR} ``continuum") \textbf{AND} TITLE-ABS-KEY (robot* \textbf{OR} ``arm" \textbf{OR} ``manipulator") \textbf{AND} TITLE-ABS-KEY (``Cosserat" \textbf{OR} ``Kirchhoff" \textbf{OR} ``Timoshenko" \textbf{OR} ``Euler-Bernoulli" \textbf{OR}  ``Geometrically exact" \textbf{OR} ``rod" \textbf{OR} ``beam") \textbf{AND} (theor* \textbf{OR} model*).
    
    \item TITLE-ABS-KEY (``soft" \textbf{OR} ``continuum") \textbf{AND} TITLE-ABS-KEY (robot* \textbf{OR} ``arm" \textbf{OR} ``manipulator") \textbf{AND} TITLE-ABS-KEY (``Cosserat" \textbf{OR} "Kirchhoff" \textbf{OR} ``Timoshenko" \textbf{OR} ``Euler-Bernoulli" \textbf{OR}  ``Geometrically exact" \textbf{OR} ``rod" \textbf{OR} ``beam") \textbf{AND} (control*).
\end{itemize}

\section{Lie Algebra} \label{sec:Appendix-B}
\begin{itemize}
\setlength \itemsep{1em}
    \item The \textit{Tilde Operator} is defined as
    
    \begin{equation}
        \tilde{\left(\cdot\right)} : \mathbb{R}^{3} \rightarrow \mathfrak{so}(3) \, .
    \end{equation}
    
    For a vector $\bm{w} = \begin{bmatrix} w_x & w_y & w_z \end{bmatrix}^{\top} \in \mathbb{R}^3$, its application gives an skew-symmetric matrix
    
    \begin{equation}
        \tilde{\bm{w}} = \begin{bmatrix} 0 & -w_z & w_y \\ w_z & 0 & -w_x \\ -w_y & w_x & 0 \end{bmatrix} \, .
    \end{equation}

    \item The \textit{Hat operator} is defined as
    
    \begin{equation}
        \hat{\left(\cdot\right)} : \mathbb{R}^{6} \rightarrow \mathfrak{se}(3) \, .
    \end{equation}
    
    Let be $\bm{h} = \begin{bmatrix} \bm{w}^{\top} & \bm{\nu}^{\top} \end{bmatrix} \in \mathbb{R}^{6}$. The \textit{Hat operator} gives
    
    \begin{equation}
        \hat{\bm{h}} = \begin{bmatrix}
                            \tilde{\bm{w}} & \bm{\nu} \\
                            \bm{0}^{\top} & 0
                        \end{bmatrix} \in \mathfrak{se}(3).
    \end{equation}

    \item The \textit{Vee operator} is the inverse of the \textit{Hat operator}
    
    \begin{equation}
        \left(\cdot\right)^{\vee} : \mathfrak{se}(3) \rightarrow \mathbb{R}^{6}.
    \end{equation}
    
    Consequentially, it is possible to write $\left(\hat{\bm{h}}\right)^{\vee} = \bm{h}$.

    \item The \textit{Adjoint Representations} consist in two maps: $\textnormal{Ad}_{\left(\cdot\right)}$ and $\textnormal{ad}_{\left(\cdot\right)}$. The former is defined as
    \begin{equation}
        \textnormal{Ad}_{\left(\cdot\right)} : SE(3) \rightarrow \mathbb{R}^{6 \times 6} , \quad
        \textnormal{Ad}_{\bm{g}} = \begin{bmatrix}
                                        \bm{R} & \bm{0}_{3 \times 3} \\ \tilde{\bm{r}} \bm{R} & \bm{R}
                                    \end{bmatrix} \, ,
    \end{equation}
    where $\bm{g} \in SE(3)$. 
    The latter is defined as
    \begin{equation}
        \textnormal{ad}_{\left(\cdot\right)} : \mathbb{R}^{6 \times 6} \rightarrow \mathbb{R}^{6 \times 6} , \quad
        \textnormal{ad}_{\bm{h}} = \begin{bmatrix}
                                        \tilde{\bm{w}} & \bm{0}_{3 \times 3} \\ \tilde{\bm{\nu}} & \tilde{\bm{w}}
                                    \end{bmatrix} \, ,
    \end{equation}
    Furthermore, it is possible to define also the co-adjoint operator $\textnormal{ad}^{*}_{\left(\cdot\right)}$, that is
    \begin{equation}
        \textnormal{ad}^{*}_{\bm{h}} = - \textnormal{ad}^{\top}_{\bm{h}}.
    \end{equation}
\end{itemize}






\bibliographystyle{IEEEtran}
\bibliography{main}

\end{document}